\documentclass[article,10pt]{IEEEtran}

\usepackage[utf8]{inputenc}
\usepackage{graphicx,cite,amsmath,amssymb,amsfonts,amsthm,mathtools,algorithm,mathabx,textcomp,blindtext,subfig,float,mwe,dblfloatfix,bm, xfrac,algpseudocode, stackengine, verbatim}
\usepackage{url}

\DeclareMathOperator*{\argmin}{argmin}
\usepackage[font=small]{caption}
\usepackage[geometry]{ifsym}
\begin{document}
	\title{Leveraging Digital Cousins for Ensemble Q-Learning in Large-Scale Wireless Networks
		\thanks{Talha Bozkus and Urbashi Mitra are with Ming Hsieh Department of Electrical and Computer Engineering, University of 
			Southern California, Los Angeles, USA. Email: \{bozkus, ubli\}@usc.edu. }
		\thanks{This work was funded by the following grants: ARO W911NF1910269, DOE DE-SC0021417, Swedish Research Council 2018-04359, NSF CCF-2008927, NSF RINGS-2148313, NSF CCF-2200221, NSF CIF-2311653, ONR 503400-78050, ONR N00014-15-1-2550 and USC + Amazon Center on Secure and Trusted Machine Learning}
	}
	
	\author{Talha Bozkus and Urbashi Mitra}

	\maketitle
\begin{abstract}
Optimizing large-scale wireless networks, including optimal resource management, power allocation, and throughput maximization, is inherently challenging due to their non-observable system dynamics and heterogeneous and complex nature. Herein, a novel ensemble $Q$-learning algorithm that addresses the performance and complexity challenges of the traditional $Q$-learning algorithm for optimizing wireless networks is presented. Ensemble learning with synthetic Markov Decision Processes is tailored to wireless networks via new models for approximating large state-space observable wireless networks. In particular, \textit{digital cousins} are proposed as an extension of the traditional digital twin concept wherein multiple $Q$-learning algorithms on multiple synthetic Markovian environments are run in parallel and their outputs are fused into a single $Q$-function. Convergence analyses of key statistics and $Q$-functions and derivations of upper bounds on the estimation bias and variance are provided. Numerical results across a variety of real-world wireless networks show that the proposed algorithm can achieve up to 50\% less average policy error with up to 40\% less runtime complexity than the state-of-the-art reinforcement learning algorithms. It is also shown that theoretical results properly predict trends in the experimental results.
\end{abstract}

\section{Introduction}\label{sec:introduction}

The optimization of large-scale real-world wireless networks is challenging due to their inherent complexity, dynamic nature, and many unobservable features \cite{wireless_network_challenges, wireless_network_challenges_2, singhal2017resource}. To overcome these challenges, digital twins have been proposed as a solution within next-generation wireless networks \cite{digital_twins_wireless_1, digital_twins_wireless_2}. In the context of wireless networks, digital twins function as virtual, exact replicas of key systems, including IoT, edge, and mobile devices, as well as network infrastructures and communication protocols. They provide a framework for the modeling, simulation, troubleshooting, and optimization of a variety of problems, such as dynamic resource allocation, interference and latency minimization, and optimal spectrum allocation \cite{dt_resource_allocation_0, dt_resource_allocation_1, dt_resource_allocation_2}. 

Herein, we use Markov Decision Processes (MDPs) to model wireless networks. MDPs provide a mathematical framework for modeling sequential decision-making problems under uncertainty \cite{bertsekas_book}. In real-world scenarios, the underlying system dynamics of wireless networks are often non-observable or difficult to estimate accurately \cite{unknown_dynamics_wireless}. To this end, we will adapt model-free reinforcement learning approaches, in particular, $Q$-learning, to optimize wireless networks. Our approach will also incorporate model estimation.

$Q$-learning \cite{bertsekas_book} has been widely employed for the optimization of wireless networks \cite{q_learning_wireless_1, q_learning_wireless_2, q_learning_wireless_3}. However, it suffers from certain limitations, particularly when dealing with large state-space wireless networks. Several variants of the $Q$-learning algorithm have been proposed to address these limitations. The estimation bias is considered in \cite{double-q, ensemble_bootstrap_q}, whereas the work of \cite{averaged_dqn, maxmin-q} deals with the estimation variance. Other methods improve learning speed \cite{speedy_q} and sample efficiency \cite{delayed_q}, deal with the high dimensionality of the state space \cite{deep_q, q_learning_func_approx}, and address exploration challenges \cite{neural_fitted_q, bootsrapped_dqn}. On the other hand, the work of \cite{q_learning_wireless_comm_1, q_learning_wireless_comm_2} is tailored to the optimization of wireless networks. These algorithms can also be categorized based on their implementation strategies, with some using a single $Q$-function estimator on a single Markovian environment \cite{bertsekas_book, speedy_q, delayed_q}, while others employ multiple $Q$-function estimators on a single Markovian environment \cite{double-q, ensemble_bootstrap_q, maxmin-q, averaged_dqn}. The design of {\bf multiple} $Q$-function estimators that can be employed on {\bf multiple} Markovian environments has not been well-studied (see \cite{talha_eusipco, talha_icassp, talha_Pn_journal}). However, the multiplicity of Markovian environments can accelerate training, improve data efficiency, and yield more accurate and stable $Q$-functions.

The traditional digital twin approach of employing exact replicas of the original Markovian system can leverage the best outcomes from different replicas (such as the most accurate or stable $Q$-functions) and enable parallel exploration of diverse states and actions. Moreover, digital twins with the same algorithms, but different parameter values, can enable more efficient parameter optimization. Despite their promising advantages, traditional digital twins have several challenges \cite{digital_twins_challenges, digital_twins_challenges_2}. They lack the ability to incorporate variations or alternative representations of the original Markovian environment, which restricts the adaptability of the algorithm to different scenarios and settings. Creating replicas that precisely mimic noisy, dynamic, or uncertain environments can enhance noise. Insufficient or biased sampling in the original environment negatively affects the accuracy of the digital twins, leading to sub-optimal learning and decision-making.

Herein, we offer the concept of \textit{digital cousins} for wireless networks in the context of reinforcement learning to overcome these challenges of traditional digital twins by exploiting our prior work \cite{talha_icassp, talha_eusipco, talha_colink_journal, talha_Pn_journal}. Digital cousins are distinct, yet structurally related synthetic Markovian environments (SMEs). These environments are inherently different, but share similar characteristics and dynamics, enabling improved performance with respect to exploration and optimization. In particular, digital cousins enable the reinforcement learning agent to explore longer trajectories, discover new state-action pairs, and learn from indirect experiences, which facilitates faster learning of the environment. These environments also help the agent exploit patterns, particularly in structured environments as found in wireless networks, and adapt to changing dynamics by leveraging past experiences. Overall, these advantages result in substantial improvements in the speed, data efficiency, accuracy, and robustness of the $Q$-learning algorithm, as will be shown later. It is worth noting that digital cousins improve exploration, and thus learning, in part because each different Markovian environment runs at a different \textit{time scale}.

Our previous work \cite{talha_colink_journal} presented low-complexity techniques for modeling and approximating observable MDPs for policy optimization. Furthermore, we emphasized the advantages of utilizing multiple SMEs to improve the accuracy and complexity of the traditional $Q$-learning algorithm across non-observable MDPs for policy learning in \cite{talha_eusipco, talha_icassp, talha_Pn_journal}. The current work  \textbf{synthesizes} both approaches and specifically focuses on the optimization of real-world wireless networks that vary in topology, scale, complexity, objective, and implementation.

The current work has key differences with respect to \cite{talha_Pn_journal}, which lead to significant improvements. Firstly, we particularly focus on the optimization of different wireless networks versus the unstructured networks considered in \cite{talha_Pn_journal}. The structured nature of wireless networks enables efficient learning by estimating the underlying model of different SMEs with lower complexity as well as the design of structural state-aggregation algorithms to minimize the memory complexity of the $Q$-learning algorithm across large state-spaces. In particular,  we employ a more accurate, robust, and practical way of modeling multiple SMEs through the use of co-link matrices; in \cite{talha_Pn_journal}, powers of the probability transition matrix are employed. The current approach captures complex, bidirectional, and interdependent dynamics of Markovian transitions within wireless networks and enhances the robustness of multiple SMEs against variability and uncertainties inherent in these transitions. 

Secondly, our proposed algorithm utilizes a straightforward policy comparison as the cost metric, avoiding the complexities of the divergence metric for $Q$-functions in \cite{talha_Pn_journal}. This design results in increased computational efficiency and enhanced resilience to noise, errors, and numerical instabilities in the $Q$-functions. These two key changes, the use of co-link matrices to model different SMEs and direct policy optimization, also yield improvements in our theoretical results. Our bounds are more accurate and require fewer assumptions, approximations, and constraints on system parameters than our prior work. Throughout the paper, we highlight the differences between the current work and our prior work \cite{talha_Pn_journal}.

Asynchronous Advantage Actor-Critic (A3C) \cite{mnih2016asynchronous} conceptually resembles our approach, Ensemble Synthetic $Q$-learning (ESQL). Both A3C and ESQL involve parallel learning to explore the state space; however, these strategies have fundamental differences. In particular, A3C is an actor-critic algorithm that models the critic (value function) and actor (policy) via neural networks. Thus, A3C suffers from the classical challenges of neural networks with respect to training time, computational complexity, robustness, and interpretability. Moreover, the critic's value function may be inaccurate or inconsistent with the actor's policy, leading to instability. ESQL utilizes $Q$-learning and hence obviates these issues.

With regards to parallel learning,  A3C uses multiple copies of the same environment (like digital twins).  In contrast, ESQL constructs synthetic environments that are \emph{distinct} but structurally related and mutually informative in a model-based fashion (digital cousins, not twins). In A3C, each network is an independent neural network initialized differently and explores the same environment in different ways, whereas, in ESQL, each network is an independent $Q$-learning learning different representations of the original environment through $n$-hop/multi-time-scale information. Unlike A3C, which periodically synchronizes network parameters, ESQL requires no synchronization; instead, $Q$-functions from each environment contribute to the global $Q$-function with weighting, enabling exploitation of the most useful environments. As ESQL exploits structural properties of multiple environments, data-efficient estimation strategies are possible to reduce sample complexity, and structural-state-action aggregation algorithms can mitigate memory complexity unlike A3C. Finally, while A3C offers performance that is competitive with ESQL, it does so with a significant increase in computational complexity; furthermore, the performance advantage of A3C for small networks degrades as the network grows in size. 

The \textbf{main contributions} of the paper are as follows:

\textbf{(i)} We leverage the innovative concept of \textit{digital cousins} as an extension of traditional digital twins and new models for approximating large state-space observable wireless networks to construct multiple distinct, yet structurally related SMEs for ensemble model-free learning. 

\textbf{(ii)} We propose a novel ensemble learning algorithm, where multiple $Q$-learning algorithms are run in parallel on multiple SMEs. Their outputs are fused into a single $Q$-function estimate via a policy comparison between different environments to produce a near-optimal deterministic policy with low complexity. 

\textbf{(iii)} We analyze the stability and convergence of the proposed algorithm from deterministic and probabilistic perspectives and provide upper-bound analysis on estimation bias and variance. We also consider a wider range of independence assumptions on the error from each digital cousin versus \cite{talha_Pn_journal}. This improved analysis is facilitated by our tailored focus on wireless networks. 

\textbf{(iv)} Numerical simulations across different real-world wireless networks show that the proposed algorithm outperforms the state-of-the-art reinforcement learning algorithms as well as methods proposed in our prior work \cite{talha_eusipco, talha_icassp, talha_Pn_journal} by achieving up to 50\% less average policy error with up to 40\% less runtime complexity. A considerable amount of reduction in memory complexity is also achieved through an original structural state-action aggregation algorithm. In addition, the experimental results are shown to follow the theoretical results closely. 

We use the following notation: vectors are bold lower
case (\textbf{x}); matrices are  bold upper case (\textbf{A}); sets are in calligraphic font ({$\mathcal{S}$}); and scalars are non-bold ($\alpha$).

\newtheorem{Proposition}{Proposition}
\newtheorem{Corollary}{Corollary}
\newtheorem{Remark}{Remark}

\section{System Model and Methods}\label{sec: mdp_model_and_methods}

\subsection{Infinite Horizon Discounted Cost MDP model}\label{subsec: infinite_horizon_model}
MDPs consist of 4-tuples $\{\mathcal{S}$, $\mathcal{A}$, $p$, $c$\}, where $\mathcal{S}$ and $\mathcal{A}$ denote the finite state and action spaces, respectively. We denote $s_{t}$ as the state and $a_{t}$ as the action taken at discrete time period $t$. The transition from state $s$ to $s^{\prime}$ under action $a$ occurs with probability $p_{a}(s,s^{\prime})$, which is stored in the $(s,s',a)^{th}$ element of the three-dimensional probability transition tensor (PTT) $\mathbf{P}$, and a bounded average cost $c_{a}(s)= \sum_{s^{\prime} \in \mathcal{S}} p_{a}\left(s, s^{\prime}\right) \hat{c}_{a}(s, s^{\prime})$ is incurred, which is stored in the $(s,a)^{th}$ element of the cost matrix $\mathbf{C}$, where $\hat{c}_{a}(s, s^{\prime})$ is the instantaneous transition cost from state $s$ to $s^{\prime}$ under action $a$. We denote the probability transition matrix (PTM) and cost vector under the action $a$ by $\mathbf{P}_a$ and $\mathbf{c}_a$, respectively. We focus on infinite horizon discounted cost MDPs, where $t = \mathbb{Z}^{+}\cup \{0\}$. Our goal is to solve \textit{Bellman's optimality} equation:  
\begin{align}
\mathbf{v}^{*}(s)&=\min _{\pi} \mathbf{v}_{\pi}(s)=\min _{\pi}\mathbb{E}_{\pi}\left[\sum_{t=0}^{\infty} \gamma^{t} c_{a_{t}}(s_{t}) | s_{0}=s\right],\label{Equ: optimization_eq}\\
\mathbf{\pi}^*(s)&=\argmin_{\pi} \mathbf{v}_{\pi}(s),\label{Equ: optimization_eq_2}
\end{align}
for all $s \in \mathcal{S}$, where $\mathbf{v}_{\pi}$ is the \textit{value function} \cite{bertsekas_book} under the \textit{policy} ${\pi}$, $\mathbf{v}^{*}$ is the \textit{optimal value function}, $\mathbf{\pi}^*$ is the \textit{optimal policy}, and $\gamma \in (0,1) $ is the discount factor. The policy $\pi$ can define either a specific action per state (\textit{deterministic}) or a distribution over the action space per state (\textit{stochastic}) for each time period. If the policy does not change over time, \textit{i.e.,}  $\pi_{t} = \pi,$ $\forall t$, then it is deemed \textit{stationary}. There always exists a deterministic stationary policy that is optimal given a finite state and action spaces \cite{bertsekas_book}. Hence, we herein consider deterministic and stationary policies.

\subsection{Model-Free Reinforcement Learning: Q-Learning}\label{subsec: Q_learning}
When the system dynamics ($p$ and $c$) are unknown or non-observable, $Q$-learning can be used to solve (\ref{Equ: optimization_eq}) and (\ref{Equ: optimization_eq_2}). The objective of $Q$-learning is to determine the optimal policy $\pi^{*}$ by learning the $Q$ functions for all state-action pairs $(s,a)$. This learning process is governed by the update rule:
\begin{equation}\label{Equ: Q-learning-update-rule}
    Q(s, a) \leftarrow(1-\alpha) Q(s, a)+\alpha(c_{a}(s)+\gamma \min _{a' \in \mathcal{A}} Q(s', a')),
\end{equation}
where $\alpha \in (0,1)$ is the learning rate. In practice, $\epsilon$\textit{-greedy} policies are used to tackle the \textit{exploration-exploitation} trade-off to ensure that sufficient sampling of the system is captured by visiting each state-action pair sufficiently many times \cite{bertsekas_book}. To this end, a random action is taken with probability $\epsilon$ (exploration), and a greedy action that minimizes the $Q$-function of the next state is taken with probability $1-\epsilon$ (exploitation). By interacting with the environment and collecting samples $\{s,a,s',c\}$, the agent updates the $Q$-functions using (\ref{Equ: Q-learning-update-rule}). The learning strategy involves determining the trajectory length ($l$) (the number of states in a trajectory), and the minimum number of visits to each state-action pair ($v$), which is generally used as a termination condition for the sampling operation. $Q$-functions converge to their optimal values with probability one, \textit{i.e.,} $Q(s,a) \xrightarrow{\mathit{w.p.1}} Q^{*}(s,a)$ for all $(s,a)$ if necessary conditions are satisfied \cite{q-learning-convergence}. The optimal policy and value functions can be derived from the $Q$-functions as:
\begin{align}
\pi^*(s)=\argmin_{a \in \mathcal{A}} Q^{*}(s, a), \hspace{10pt} \mathbf{v}^*(s)=\min_{a \in \mathcal{A}} Q^{*}(s, a).
\end{align}

\setlength{\textfloatsep}{8pt}
\begin{figure}[t]
    \centering
    \includegraphics[width=0.45\textwidth]{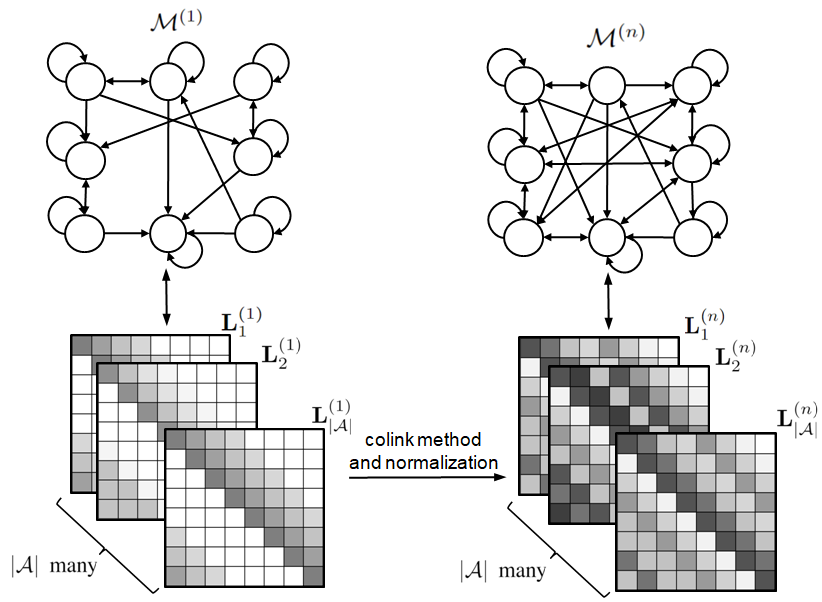}
    \caption{The relationship between $\mathbf{L}^{(1)}$ (= $\mathbf{\hat{P}}$), $\mathbf{L}^{(n)}$, $\mathcal{M}^{(1)}$ and $\mathcal{M}^{(n)}$}
    \label{Fig:relation_between_M1_and_Mn}
\end{figure} 

\subsection{Extension of Digital Twins: Digital Cousins}\label{subsec: digital_cousins}

In our prior work \cite{talha_eusipco, talha_icassp, talha_Pn_journal}, we introduced the novel concept of $\textit{digital cousins}$ for general networks, which represents a set of closely related synthetic Markovian environments that share analogous structures and characteristics with both the original Markovian environment and each other. This approach effectively captures the essence of the original Markovian environment while introducing variations, addressing many of the limitations of traditional digital twins.

There are several ways to create SMEs (and the corresponding PTTs) based on the PTT of the original environment $\mathbf{P}$. A natural approach is to employ a function of $\mathbf{P}$ or $\mathbf{P}^T$. It is desirable to maintain a consistent relationship between transition probabilities $p_a(s, s')$ across different environments while ensuring that the PTTs of different environments are row-stochastic or can be converted to such form via appropriate normalization techniques without altering the underlying structures. We herein employ the $\textit{colink method}$, which was originally introduced in \cite{co_link_paper}, which involves constructing a set of symmetric matrices based on the original probability transition matrix. Our prior work \cite{talha_colink_journal} demonstrated the effectiveness of the colink method for policy optimization in large state-space wireless networks (with known $\mathbf{P}$). Herein, the structure of wireless networks can be exploited to improve the estimation of the co-link matrices for learning in contrast to the unstructured network models considered in \cite{talha_Pn_journal}. 

The colink method produces the \textit{$n^{th}$ order similarity tensor}, denoted by $\mathbf{L}^{(n)}$ ($\mathbf{L}^{(n)}_i$ is the matrix for the $i^{th}$ action), which can be expressed as: 
\begin{equation}\label{Equ: sum_expression_for_Ln}
    \mathbf{L}^{(n)}_i = \sum_{k = 0}^{n-2} \mathbf{P}^{n-k-1}_i(\mathbf{P}^{T}_i)^{k+1} + (\mathbf{P}^{T}_i)^{n-k-1}\mathbf{P}^{k+1}_i, \hspace{4pt} \forall i \in \mathcal{A},
\end{equation}
where $\mathbf{P}_i$ is the PTM corresponding to the $i^{th}$ action in $\mathbf{P}$. We employ $l_1$ normalization on $\mathbf{L}^{(n)}_i$, $\forall i \in \mathcal{A}$ (so that the sum of each row in $\mathbf{L}^{(n)}_i$ is 1). This transformation preserves the structure of $\mathbf{L}^{(n)}$ while generating row-stochastic PTMs, as discussed in \cite{talha_colink_journal}. The resulting synthetic Markovian environment, denoted as $\mathcal{M}^{(n)}$, captures the $n$-step probabilistic transitions between states represented by $\mathbf{L}^{(n)}$.

When dealing with large dimensions of $\mathbf{P}$ or a high order $n$, utilizing (\ref{Equ: sum_expression_for_Ln}) can pose computational challenges due to the increasing number of matrix multiplications involved. However, it is possible to overcome this issue by leveraging the structural properties of $\mathbf{P}$. By exploiting properties such as diagonal, circulant, Toeplitz, or block structures, alternative approaches can be developed to compute (\ref{Equ: sum_expression_for_Ln}) with significantly lower complexity \cite{talha_colink_journal}.

The original PTT, $\mathbf{P}$, is initially unknown; thus, it needs to be estimated to create the PTTs of multiple SMEs. If the number of one-step transitions between states can be counted, the transition probabilities can be accurately estimated via \textit{sample averaging} \cite{sample_averaging_2}. If there is a particular structure in one-step transition probabilities, as found in wireless networks, a more data-efficient estimation can also be performed \cite{talha_colink_journal, talha_eusipco}. We denote the estimated PTT by $\mathbf{\hat{P}}$, and construct $\mathbf{L}^{(n)}$ using $\mathbf{\hat{P}}$. The relationship between the original Markovian environment ($\mathcal{M}^{(1)}$) and the $n^{th}$ synthetic Markovian environment ($\mathcal{M}^{(n)}$, $n>1$) is given in Fig.\ref{Fig:relation_between_M1_and_Mn}.

The accuracy of $\mathbf{L}^{(n)}$ and $\mathcal{M}^{(n)}$ is affected by the quality of sampling and the order $n$. Insufficient sampling can lead to accumulated errors during matrix multiplications, resulting in lower accuracy for higher-order environments ($\textit{i.e.}$ for large $n$). Additionally, as $n$ increases, the interpretation of $\mathbf{L}^{(n)}$ becomes less intuitive, making it challenging to understand state transitions over longer time horizons. While $\mathbf{L}^{(n)}$ tends to converge to a fixed tensor as $n$ increases, this convergence may also cause the loss of underlying system dynamics. In particular, our prior work \cite{talha_colink_journal} showed the existence of an optimal $n$ that maximizes the accuracy of the estimated policy. We underline that this optimal $n$ is not large, of the order 5.

The use of colink methods provides several advantages over our prior approach \cite{talha_Pn_journal}. Co-link modeling effectively captures \textbf{bidirectional} relationships in Markovian transitions, which enables improved modeling of feedback loops and other interdependencies seen in wireless networks\cite{talha_colink_journal}. Incorporating bidirectional relationships can also enhance the robustness of the synthetic Markovian environments against stochastic impairments that might be present in the underlying Markovian transitions. The nature of colink representations also ensures sufficient variability across multiple SMEs. 

\section{Algorithm}\label{sec: link_learning_algorithm}
\begin{figure}[t]
    \centering
    \subfloat[Original QL]{{\includegraphics[width=1.84cm]{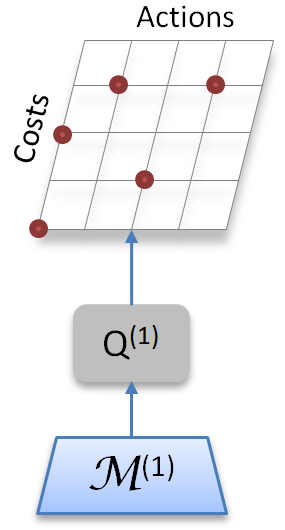}}}%
    \hspace{2pt}
    \subfloat[Ensemble QL]{{\includegraphics[width=2.95cm]{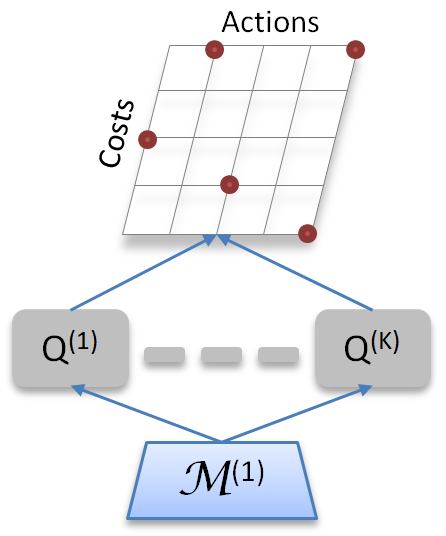}}}%
    \hspace{2pt}
    \subfloat[Proposed QL]{{\includegraphics[width=3.5cm]{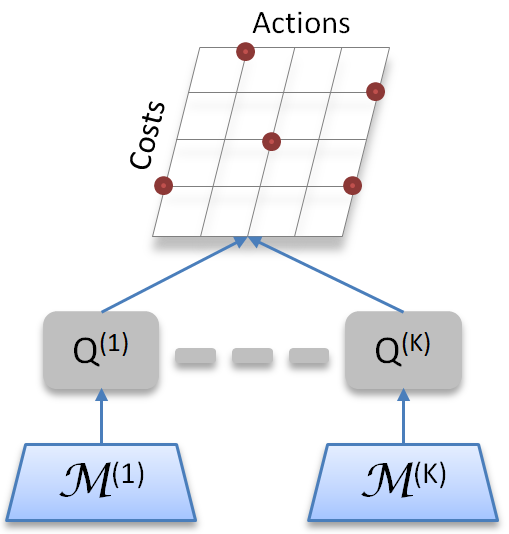}}}%
    \caption{Comparison of Q-Learning (QL) algorithms based on their implementation strategies}
    \label{fig:different_ql_algo}
\end{figure}

This section introduces the proposed algorithm, Ensemble Synthetic $Q$-Learning (ESQL) (Algorithm \ref{Algorithm: ensemble_link_learning}). We leverage $K-1$ SMEs ($\mathcal{M}^{(n)}$ for $n \in \{2,...,K\}$), in addition to the original Markovian environment $\mathcal{M}^{(1)}$, resulting in a total of $K$ Markovian environments. The comparison between the traditional $Q$-learning algorithm, conventional ensemble $Q$-learning algorithms, and the proposed algorithm is given in Fig.\ref{fig:different_ql_algo}, where $\mathbf{Q}^{(n)}$ represents the $Q$-function estimator for the $Q$-learning algorithm running on $\mathcal{M}^{(n)}$ for $n \in \{1,2,...,K\}$.

Algorithm \ref{Algorithm: ensemble_link_learning} has several inputs: the trajectory length ($l$), the minimum number of visits required for each state-action pair ($v$), the update ratio at time $t$ ($u_t \in [0,1]$), the total number of Markovian environments ($K$), and the empty $Q$-tables for the $K$ different environments ($\mathbf{Q}^{(n)}$ for $n \in \{1,2,...,K\}$). Let $\mathbf{w}_t$ denote the weight vector of size $K$ at time $t$, where $\textbf{w}^{(n)}_t$ represents the $n^{th}$ element of $\mathbf{w}_t$. At $t=0$, the weight vector $\mathbf{w}_0$ is initialized randomly in line 1. Each element is chosen uniformly random from the range [0,1], and the vector is subsequently softmax-normalized to ensure that $\sum_{n=1}^K \textbf{w}^{(n)}_0 = 1$. This initialization serves to break the symmetry. The iterations continue until each state-action pair in $\mathcal{M}^{(1)}$ is visited at least $v$ times (in line 2), ensuring sufficient capture of different state-action dynamics. At the end of each trajectory ($\textit{i.e.}$ every $l$ time step), all $K$ Markovian environments are reset, and a common initial state is assigned randomly from $\{1,2,...,|\mathcal{S}|\}$, as indicated in line 3. Independent samples are collected from each different Markovian environment, and corresponding $Q$-tables are updated independently in line 6. We note that although the initial states are the same, different actions are taken following the epsilon-greedy policy of each different environment; thus, different next-state and cost pairs are observed for different environments. This procedure is repeated $l$ times, after which a random but common initial state is set. 

In line 7, we obtain a set of \textit{individual policies}, $\bm{\pi}_t^{(n)}$, by minimizing the current $Q$-functions $\mathbf{Q}_t^{(n)}$. In line 8, we compute the \textit{correct estimation rate vector} $\mathbf{w}_t$, and its $n^{th}$ element ($\mathbf{w}_t^{(n)}$) represents how close the individual policy $\bm{\pi}_t^{(n)}$ is to the policy of the original environment $\bm{\pi}_t^{(1)}$. Hence, $\mathbf{w}_t^{(n)}$ is a good indicator of how useful the most recent samples from $\mathcal{M}^{(n)}$ are for constructing the ensemble output. In line 10, the vector $\mathbf{w}_t$ is $\mbox{softmax}$-normalized, and used to update the $Q$-function output of Algorithm \ref{Algorithm: ensemble_link_learning}, denoted as $\mathbf{Q}^{it}_t$, in line 11. During the update of $\mathbf{Q}^{it}_t$, the algorithm balances \textit{exploitation} by utilizing a fraction $u_t$ of the $\mathbf{Q}^{it}_t$ from the previous iteration, while promoting \textit{exploration} by sampling multiple Markovian environments based on their weights, with their contributions weighted by $1-u_t$. The estimated policy $\hat{\bm{\pi}}$ is derived from $\mathbf{Q}^{it}$ in line 15. 

Our approach directly optimizes \textbf{policies} and compares the optimal policies of different environments to determine weights (in lines 7-8) in contrast to \cite{talha_Pn_journal}, where the $Q$-functions are optimized and utilized in the weighting mechanism (see Algorithm 1 of \cite{talha_Pn_journal}, lines 7-8). This approach simplifies algorithm design and improves performance in several ways. Firstly, policy comparison is computationally efficient, as it only involves comparing selected actions instead of the entire $Q$-table, resulting in faster computation. Secondly, updating the policy directly robustifies against bias, error, or noise in the $Q$-functions. Thirdly, policies are represented by discrete actions, which eliminates issues associated with taking the difference between small $Q$-functions, thus providing numerical stability. Lastly, direct policy optimization yields more interpretable results.

The \textit{iterative} update of $\mathbf{Q}^{it}_t$ with the current weight vector $\mathbf{w}_t$ captures \textit{asymmetric} information between different Markovian environments. Initially, $\mathcal{M}^{(1)}$ provides more useful samples as it is the original Markovian environment, and there are limited observations for higher-order relationships versus the first-order relationships. As iterations progress, $\mathcal{M}^{(n)}$ for larger $n$ contributes more, improving exploration and reducing reliance on the first-order relationships.

\algdef{SE}[REPEATN]{RepeatN}{End}[1]{\algorithmicrepeat\ #1 \textbf{times}}{\algorithmicend}
\begin{algorithm}[t]
\caption{Ensemble Synthetic $Q$-Learning (ESQL)}
\hspace*{\algorithmicindent} \textbf{Input:} $l, v, u_t, K$, $\mathbf{Q}^{(n)}, n \in \{1,2,...,K\}$ \\
\hspace*{\algorithmicindent} \textbf{Output: $\mathbf{Q}^{it}$, $\hat{\bm{\pi}}$} 
\begin{algorithmic}[1]
    \State Initialize $\mathbf{w}_0$ randomly, $\mathbf{Q}^{it}_0 \gets \mathbf{0}$, $t \gets 0$
    \While{each $(s,a)$ pair in $\mathcal{M}^{(1)}$ not visited $v$ times}
        \State choose common initial state for all $\mathcal{M}^{(n)}$ randomly
        \RepeatN{$l$}
            \For{each $n \in \{1,...,K\}$}
                \State sample $\{s,a,s',c\}$ from $\mathcal{M}^{(n)}$ and update $\mathbf{Q}^{(n)}_t$ using (\ref{Equ: Q-learning-update-rule})
                \State $\bm{\pi}_t^{(n)}(s) \gets \operatorname*{argmin}_{a'} \mathbf{Q}^{(n)}_t(s,a')$
                \State $\mathbf{w}^{(n)}_t \gets \frac{1}{|\mathcal{S}|} \sum_{j=1}^{|\mathcal{S}|} \mathbf{1}(\bm{\pi}^{(1)}_t(j) = \bm{\pi}^{(n)}_t(j))$
                \EndFor
            \State $\mathbf{w}_t \gets \mbox{softmax}(\mathbf{w}_t)$
            \State $\mathbf{Q}^{it}_{t+1} \gets u_t\mathbf{Q}^{it}_t+(1-u_t)\sum_{n=1}^{K}\mathbf{w}^{(n)}_t\mathbf{Q}^{(n)}_t$
            \State $t \gets t+1$
        \End
    \EndWhile
    \State $\hat{\bm{\pi}}(s) \gets \operatorname*{argmin}_{a'}\mathbf{Q}^{it}(s,a')$
\end{algorithmic}
\label{Algorithm: ensemble_link_learning}
\end{algorithm}

We emphasize that our proposed algorithm combines the features of online and offline RL methods. In real-time, we estimate the PTT of the original environment through sampling and update the corresponding $Q$-functions (\textbf{online} part). Simultaneously, we construct PTTs for multiple SMEs and learn their $Q$-functions using pre-collected data (\textbf{offline} part). Unlike hybrid RL algorithms \cite{hybrid_rl}, which uses both offline and online data collection, our approach continuously updates $Q$-functions in real-time while leveraging pre-collected data for improved accuracy and stability.

While our approach includes model-based learning components such as estimating the PTT of the original Markovian environment, it differs from model-based reinforcement learning \cite{model_based_RL} in several ways: (i) We focus on learning $Q$-functions using $Q$-learning across multiple Markovian environments to determine optimal policies, rather than accurately modeling environment dynamics. (ii) The estimated PTT is used to generate multiple SMEs to facilitate learning and exploration, and is not directly used for decision-making. (iii) We focus on sampling-driven cost collection as opposed to explicit cost function estimation.

\subsection{Deterministic analysis}
\label{subsec:deterministic_theory}

We next present a set of theoretical results for Algorithm \ref{Algorithm: ensemble_link_learning}, including convergence results and stability analysis in a deterministic setting. We underscore that in \cite{talha_Pn_journal}, the analysis is mostly for the case of a distributional assumption on the $Q$-function errors. We shall tackle the probabilistic analysis for our current framework in the sequel. In addition to a series of results for the deterministic case; our analysis herein has some key differences from that in \cite{talha_Pn_journal}. As previously noted, the incorporation of wireless networks and the use of co-link representations for the network dynamics enables us to exploit structures, facilitating tighter results. Additionally, given the consideration of direct policy comparison versus a divergence error metric for the $Q$-functions in \cite{talha_Pn_journal}, there are additional simplifications and tighter results possible. Finally,  we observe that several results in \cite{talha_Pn_journal} require assumptions and approximations. An example assumption is needing a particular time-varying update ratio to ensure convergence, whereas, herein, we can employ a constant update ratio $u_t = u$.

\begin{Proposition}\label{Proposition-1}\normalfont 
Let $\Delta_{t}^{it}(s,a)$ denote the $Q$-function update of the output of Algorithm \ref{Algorithm: ensemble_link_learning} for $(s,a)$ from time $t$ to $t+1$ as: $\Delta_{t}^{it}(s,a) = \mathbf{Q}_{t+1}^{it}(s,a) - \mathbf{Q}_{t}^{it}(s,a)$.  Then, 
\begin{align}
    \lim_{t\rightarrow\infty}|\Delta_{t}^{it}(s,a) - \Delta_{t-1}^{it}(s,a)| = 0,
\end{align}
for all $(s,a)$. (See Appendix \ref{Appendix: proposition_1})
\end{Proposition}

This proposition provides insight into the behavior of the $Q$-function updates over time. In particular, as more iterations are performed, the updates to the $Q$-function become increasingly stable, indicating the stabilization of the overall learning process. Our result leverages the convergence properties of the traditional $Q$-learning \cite{bertsekas_book} and the convergence of the weights. We note that Proposition 3 \cite{talha_Pn_journal}  bounds the difference between the $Q$-functions of the original Markovian environment and different SMEs; while Proposition 1 herein shows that the ensemble $Q$-function converges to a fixed $Q$-function, and consequently, the ensemble policy converges to a fixed policy.

\begin{Proposition}\label{Proposition-2}\normalfont
Let $\epsilon_{t}^{(n)}(s,a)$ be the weighted $Q$-function update of the $n^{th}$ environment for $(s,a)$ from time $t$ to $t+1$ as: $\epsilon_{t}^{(n)}(s,a) = \mathbf{w}_{t+1}^{(n)} \mathbf{Q}_{t+1}^{(n)}(s,a)  - \mathbf{w}_{t}^{(n)} \mathbf{Q}_{t}^{(n)}(s,a)$. Moreover, let $\theta^{(n)}(s,a)$ be the smallest positive constant that satisfies $|\epsilon^{(n)}_{t}(s,a)| \leq \theta^{(n)}(s,a)$ for all $t, n, (s,a)$. Then, 
\begin{align}\label{Equ: proposition_2}
    \lim_{t\rightarrow\infty}|\Delta_{t}^{it}(s,a)| \leq \sum_{n=1}^{K}\theta^{(n)}(s,a),
\end{align}
for all $(s,a)$. (See Appendix \ref{Appendix: proposition_2})
\end{Proposition}

The proposition provides an upper bound on $|\Delta_{t}^{it}(s,a)|$ in the limit depending on the characteristics of the weighted $Q$-function updates in each environment. The environments with larger $\theta^{(n)}$ values will have a greater impact on the bound. Furthermore, a tighter upper bound indicates more stable and controlled updates, potentially leading to faster convergence and improved learning efficiency. 

Proposition 2 bounds the summed error over each Markovian environment and thus, this bound is implicitly a function of the number of Markovian environments, $K$. In contrast, Proposition 2 \cite{talha_Pn_journal} shows that the scaling of the error variance between the ensemble $Q$-function and the optimal one is bounded by a constant that scales inversely with $K$. It should be noted that both propositions directly exploit the bounded nature of key functions albeit rather different ones: weighted $Q$-function errors and update ratio $u$ here and the weights and the $\mbox{softmax}$ operator in \cite{talha_Pn_journal}.

\begin{Corollary}\normalfont
It takes at most $t = \frac{\log(1 - \frac{\beta}{\sum_{n=1}^{K}\theta^{(n)}(s,a)})}{\log(u)}$ iterations to ensure that $|\Delta_{t}^{it}(s,a)|\leq\beta$ for any $\beta>0$ and $(s,a)$. (See Appendix \ref{Appendix: corollary_1})
\end{Corollary}

By appropriately choosing the parameter $u$ in Algorithm \ref{Algorithm: ensemble_link_learning}, we can effectively control the speed of convergence and the accuracy of the $Q$-function updates. In particular, a larger $u$ leads to faster convergence.

\begin{Proposition}\label{Proposition-3}\normalfont 
If there exist positive constants $\phi^{(n)}(s,a) \in (0,1)$ such that $\frac{|\epsilon^{(n)}_{t+1}(s,a)|}{|\epsilon^{(n)}_{t}(s,a)|}$ $\leq$ $\phi^{(n)}(s,a)$ for all $n,t,(s,a)$, then
\begin{align}
    \lim_{t\rightarrow\infty}|\Delta_{t}^{it}(s,a)| = 0
\end{align}
for all $(s,a)$. (See Appendix \ref{Appendix: proposition_3})
\end{Proposition}

This proposition provides a sufficient deterministic condition for the convergence of Algorithm \ref{Algorithm: ensemble_link_learning}. If the weighted $Q$-function update of each Markovian environment is non-increasing over time, Algorithm \ref{Algorithm: ensemble_link_learning} converges to the optimal $Q$-functions. We will show numerically for different settings that the constants $\phi^{(n)}(s,a)$ exist with high probability. We establish this result by utilizing the non-increasing nature of weighted $Q$-function errors under the assumption of constant $u_t$, while Corollary 2 of \cite{talha_Pn_journal} relies on a strict time-varying form on the update ratio of the algorithm and independence assumptions on the stochastic $Q$-function errors.

\begin{Proposition}\label{Proposition-4}\normalfont The weights $\mathbf{w}^{(n)}_t$ in Algorithm \ref{Algorithm: ensemble_link_learning} converge to the following final weights:
\begin{align}
    {\mathbf{w}^{(n)}} = \frac{e^{\frac{1}{\mathcal{S}}\sum_{j=1}^{|\mathcal{S}|} \mathbf{1}(\bm{\pi}^{(1)}(j) = \bm{\pi}^{(n)}(j))}}{\sum_{i=1}^K e^{\frac{1}{\mathcal{S}}\sum_{j=1}^{|\mathcal{S}|} \mathbf{1}(\bm{\pi}^{(1)}(j) = \bm{\pi}^{(i)}(j))}}, \label{Equ: new_weights_formula}
\end{align}
where $\bm{\pi}^{(n)} = \underset{\bm{\mu}}{\arg\min}(\mathbf{I} - \gamma\mathbf{\tilde{L}}^{(n)}_{\bm{\mu}})^{-1}\mathbf{c}_{\bm{\mu}}$ and $\mathbf{\tilde{L}}^{(n)}_{\bm{\mu}}$ is the $l_1$-normalized version of $\mathbf{L}^{(n)}_{\bm{\mu}}$ for all $n$.
\end{Proposition}

This proposition shows that the final weights of Algorithm \ref{Algorithm: ensemble_link_learning} can be computed without relying on the $Q$-functions. Furthermore, the final weights are non-monotonic across the order $n$, but the weight for the original system ($n=1$) is always the largest. We note that the order of weights can change across iterations, as will be shown numerically. This result can be easily shown using the compact expression for the $Q$-functions \cite{bertsekas_book} and the fact that the weight vector $\mathbf{w}$ is normalized using the $\mbox{softmax}$ operator in Algorithm \ref{Algorithm: ensemble_link_learning}. We herein have a closed-form solution to the convergent weights of Algorithm \ref{Algorithm: ensemble_link_learning}; thus, one can directly compare the relative importance of each Markovian environment. In contrast, Proposition 4 \cite{talha_Pn_journal} provides a partial ordering between the $Q$-functions of any Markovian environment; thus, the results are less precise.

\subsection{Probabilistic analysis}
\label{subsec:probabilistic_theory}
The results above did not make any assumptions about the underlying distributions of the $Q$-function \textbf{errors}. Herein, as in \cite{talha_Pn_journal}, we assume that the $Q$-function errors follow a distribution.  In particular, the distribution $D$ has zero-mean and finite variance as follows:
\begin{align}
    \mathcal{X}^{(n)}_t(s,a) = \mathbf{Q}_{t}^{(n)}(s,a) - \mathbf{Q}^{*}(s,a)  \sim D\Big(0,\frac{\lambda_n^2}{3}\Big), \label{Equ: distribution_assumption}
\end{align}
for all $n, (s,a)$ with $\lambda_n > 0$ where $\mathbf{Q}^{*}$ is the optimal $Q$-functions of the original Markovian environment. Unlike prior studies that assumed specific distributions for $D$ \cite{uniform_assump_1}, we take a more general approach by adopting a common distribution family that governs the $Q$-function errors across all environments, as employed and validated in \cite{talha_icassp, talha_Pn_journal}. This assumption is reasonable due to the shared state and action space, the use of the same cost function, and common learning parameters across all Markovian environments. Simulations also confirm that the true distribution $D$ exhibits a zero-mean and finite variance for all $n$ (see Fig.~\ref{fig:q_func_error_all_n}). A key difference between our current work and our prior work \cite{talha_Pn_journal} is that we herein consider a variety of different independence assumptions on the errors between different Markovian environments, which capture the dynamics of different wireless network environments more effectively relative to \cite{talha_Pn_journal}.

Let $\mathbb{E}$ and $\mathbb{V}$ be the expectation and variance operators, $\lambda = \max\limits_{n \in \{1,2,...,K\}}\lambda_n$, and $\mathcal{E}_t(s,a) = \mathbf{Q}_{t}^{it}(s,a) - \mathbf{Q}^{*}(s,a)$.

\begin{Proposition}\label{Proposition-5}\normalfont
Let $u_t$ be constant: $u_t=u$. Under assumption (\ref{Equ: distribution_assumption}), Algorithm \ref{Algorithm: ensemble_link_learning} produces unbiased $Q$-functions in the limit $\textit{i.e.}$ $\lim\limits_{t\rightarrow \infty}\mathbb{E}[\mathcal{E}_t(s,a)] = 0$ for all $(s,a)$, and the upper bounds on the error variance in the limit based on different independence assumptions are given in Table \ref{Table:upper_bounds}. (See Appendix \ref{Appendix: proposition_5})
\end{Proposition}

\begin{table}[t]
\centering
\small
\setlength{\tabcolsep}{5pt}
\begin{tabular}{|c| c|} 
 \hline
 Independence assumption & Upper bound \\ [1ex]
 \hline
  $\mathcal{X}^{(n)}_t \perp \mathcal{X}^{(m)}_t$ $\forall t$, $\mathcal{X}^{(n)}_{t_1} \perp \mathcal{X}^{(n)}_{t_2}$ $\forall n$ (strict) &  $\frac{(1-u)}{(1+u)}\frac{\lambda^2}{3}$ \\ [1ex]
 \hline
 $\mathcal{X}^{(n)}_{t_1} \perp \mathcal{X}^{(n)}_{t_2}$ $\forall n$ (modest) & $\frac{(1-u)}{(1+u)}\lambda^2$ \\ [1ex]
 \hline
 no independence assumption & $\frac{2\lambda^2}{(1+u)^2}+\frac{(1-u)}{(1+u)}\lambda^2$ \\ [1ex]
 \hline
\end{tabular}
\caption{Upper bounds on $\lim\limits_{t\rightarrow \infty}\mathbb{V}[\mathcal{E}_t(s,a)]$}
\label{Table:upper_bounds}
\end{table}

This proposition shows that under the assumption (\ref{Equ: distribution_assumption}), $\mathbf{Q}_{t}^{it}$ is an unbiased estimator of $\mathbf{Q}^{*}$ in the limit, and the variance of the error can always be upper bounded. Clearly, the stricter independence assumption leads to a tighter bound. Herein, a larger $\lambda$ implies a higher uncertainty in the $Q$-function errors, yielding looser upper bounds. On the other hand, when the algorithm converges ($t \rightarrow \infty$), a larger $u$ leads to less reliance on SMEs. We note that when there is high uncertainty in the system, upon initialization, the digital cousins provide observations at multiple time scales, improving exploration. However, it should be underscored that the optimal $Q$-functions for each SME may not be the same, and be different from that of the original Markovian environment. Thus, as uncertainty diminishes, we want a lower reliance on SMEs and, thus, a larger $u$.

\begin{Corollary}\normalfont\label{Corollary_2}
By assuming a specific time-varying structure on the parameter $u_t$, as stated in Proposition 5, and an independence assumption on the $Q$-function errors, it is possible to derive a sufficient probabilistic condition for convergence for Algorithm \ref{Algorithm: ensemble_link_learning}. This condition can be employed to show convergence when Proposition \ref{Proposition-3} fails to apply \cite{talha_icassp}.
\end{Corollary}

\section{Numerical Results}\label{sec: numerical_results}

In this section, we evaluate the accuracy and complexity of the proposed algorithm in wireless networks, considering four distinct wireless network models. These models vary in their topology, scale, complexity, objective, and implementations and serve as representative examples that accurately depict the complexities and challenges of real-world wireless networks. The detailed descriptions and parameter optimizations can be found in \cite{Ln_journal_supplementary_material}. We assume that data arrivals to each transmitter in the first three network models follow an independent and identically distributed distribution, such as Bernoulli as in \cite{talha_colink_journal}.

\subsection{Wireless Network Models}\label{subsec: wireless_network_models}

\subsubsection{MISO network with interference channels} 

There are multiple transmitters (TX$_1$, TX$_2$, TX$_3$) and a single receiver (RX) as shown in Fig.\ref{fig:miso_collision_network}. Each transmitter has a finite-length buffer that stores the incoming data packets, and the channel between each transmitter and receiver is modeled with a multi-state Gilbert-Elliot channel model \cite{talha_colink_journal} with different parameters. Each transmitter experiences its own channel to the receiver; furthermore, collisions can occur due to interference between transmitters transmitting simultaneously. The magnitude of interference is inversely proportional to the distance between transmitters. The objective is to determine when each transmitter should \textit{transmit} data or \textit{remain silent} given the buffer load of each transmitter and overall channel conditions in order to minimize the overall cost. The cost function has three components: the buffer cost, the channel cost (incurred when the transmitter transmits under unfavorable channel conditions), and the collision cost (proportional to the number of data packets being transmitted simultaneously).

\subsubsection{MISO energy harvesting network with multiple relays} 

\begin{figure}[t]
    \centering
    \subfloat[Model 1: MISO network with interference channels]{{\includegraphics[width=6.2cm]{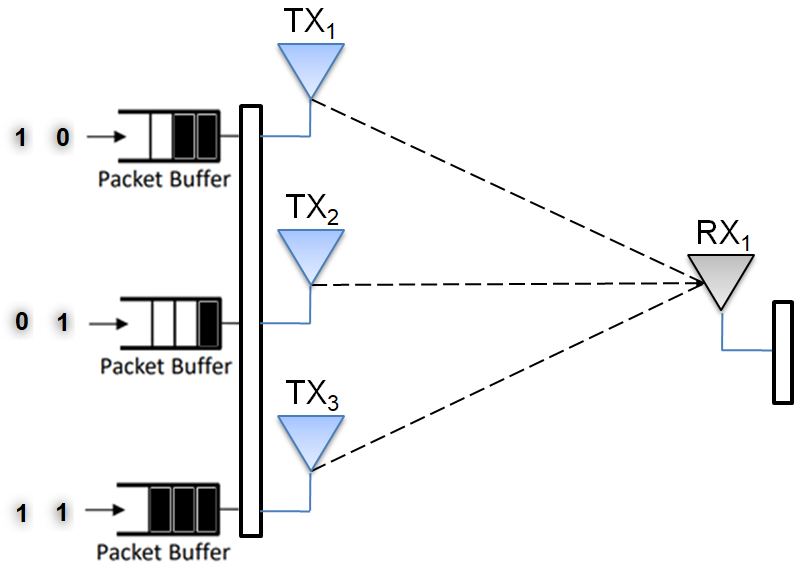}}\label{fig:miso_collision_network}}%
    
    \subfloat[Model 2: MISO energy harvesting network with Gaussian interference channels and multiple relays]{{\includegraphics[width=6.5cm]{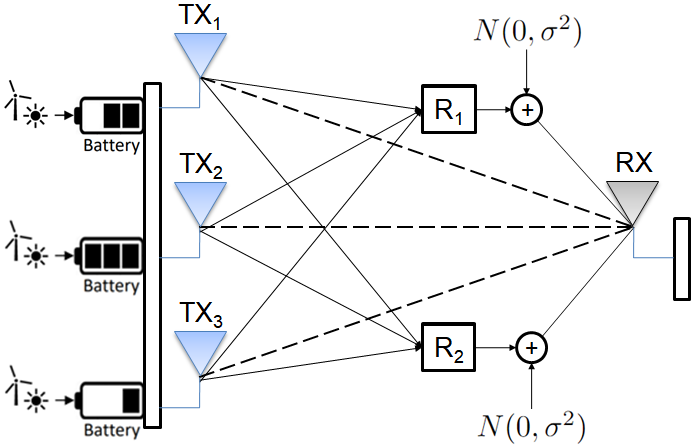}}\label{fig:miso_relays_network}}%
    
    \subfloat[Model 3: MIMO network with interference channels ]{{\includegraphics[width=6.3cm]{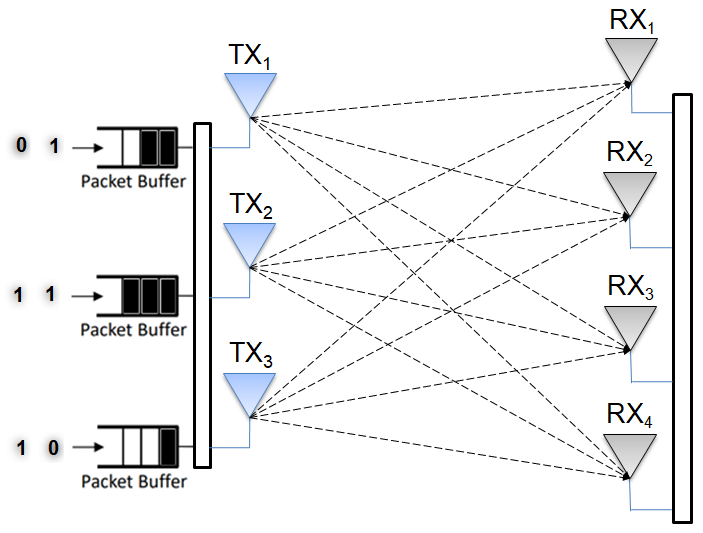}}\label{fig:mimo}}%
    
    \subfloat[Model 4: MIMO network with mobile transmitters]{{\includegraphics[width=7.5cm]{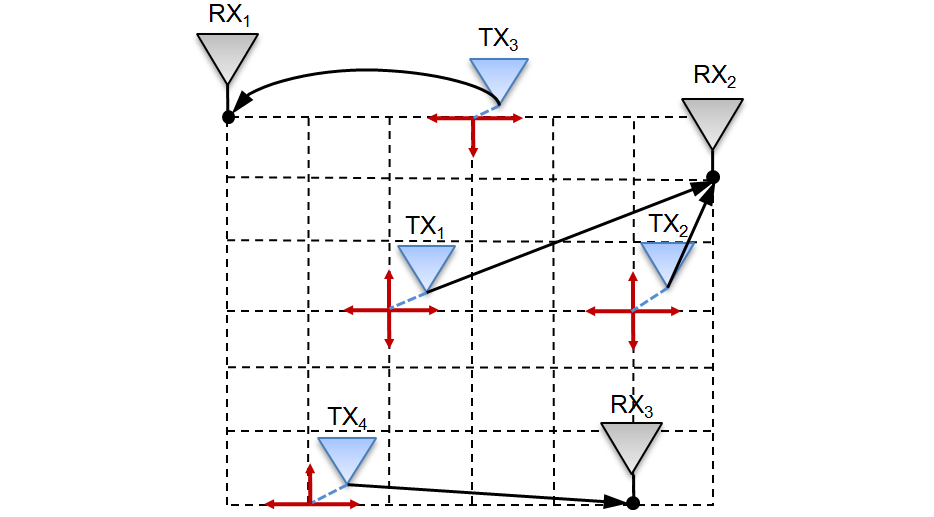}}\label{fig:moving_case}}%
    \caption{Examples of wireless network models.}
\end{figure}

There are multiple transmitters (TX$_1$, TX$_2$, TX$_3$), multiple relays (R$_1$, R$_2$), and a single receiver (RX) as shown in Fig.\ref{fig:miso_relays_network}. Each transmitter is equipped with a finite-sized battery that stores the harvested energy packets in discrete units. All the channels are modeled with a standard Gilbert-Elliot model, and the channel parameters are chosen such that the channel between transmitters and relays as well as relays and receivers are more likely to be in a good state than the direct channel between transmitters and receivers (due to shorter distance and less interference). The outputs of the relays are corrupted by AWGN noise. The objective is to determine when transmitters should \textit{directly} transmit or transmit \textit{through relays} given the battery load of each transmitter and overall channel conditions in order to minimize the overall cost. The cost function consists of three components: the negative throughput (inversely proportional to successfully received packets), the drop cost (balancing relays' load), and the battery consumption cost (proportional to total battery usage).

\subsubsection{MIMO network with interference channels} 
We consider a MIMO network model with multiple transmitters (TX$_1$, TX$_2$, TX$_3$), and multiple receivers (RX$_1$, RX$_2$, RX$_3$, RX$_4$) as shown in Fig.\ref{fig:mimo}. Each transmitter is equipped with a finite-sized data buffer for storing incoming data packets. All the channels are modeled with a standard Gilbert-Elliot model, and the channels between closer TX-RX pairs (such as TX$_1$ and RX$_1$) are more likely to be in a good state compared to the channels between farther TX-RX pairs (such as TX$_1$ and RX$_4$) due to shorter distance and less interference. Each channel allows only one data transmission at a time. If a transmitter needs to send data to multiple receivers, the channels in the best conditions will be used to optimize transmission. Transmitters sending packets to the same receiver cause interference with each other. The magnitude of interference is inversely proportional to the distance between the transmitters. The objective is to determine \textit{how many data packets} each transmitter should transmit given the buffer load of each transmitter and overall channel conditions in order to minimize the overall cost. The cost function comprises four components: the buffer cost, the channel cost, the collision cost, and the receiver load cost (balancing receivers' load).

\subsubsection{MIMO network with mobile transmitters} 
We consider a MIMO network model with multiple mobile transmitters (TX$_1$, TX$_2$, TX$_3$, TX$_4$) and multiple fixed receivers (RX$_1$, RX$_2$, RX$_3$) as shown in Fig.\ref{fig:moving_case}. Each transmitter moves randomly at different constant speeds within possible directions, as indicated by the red arrows. Their movement is constrained within a specified bounded area. They can only move in integer multiples of unit steps and are not allowed to occupy the same location as receivers. The channel between the pair of transmitters and receivers follows a path-loss model. The objective is to determine \textit{the optimal association between transmitters and receivers} given the speed of transmitters, the positions of transmitters and receivers, and the overall channel conditions in order to minimize the overall cost. The black arrows illustrate one possible association. The cost function comprises three components: the negative throughput, the receiver load cost, and the interference cost (inversely proportional to the squared distance between transmitters transmitting to the same receiver). 

\begin{figure}[t]
    \centering
    \includegraphics[width=0.35\textwidth]{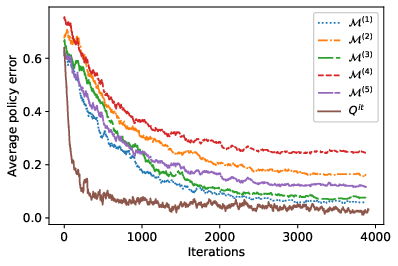}
    \caption{APE performances across different environments}
    \label{Fig:sharp_decrease}
\end{figure} 

\subsection{Average Policy Error (APE) Performance}\label{subsec: APE_results}

Let $\bm{\pi^{*}}$ be the optimal policy from (\ref{Equ: optimization_eq_2}), and $\hat{\bm{\pi}}$ be the output policy of Algorithm \ref{Algorithm: ensemble_link_learning}. We define the \textit{average policy error (APE)} as follows:
\begin{align}
    APE &=\frac{1}{|\mathcal{S}|} \sum_{s=1}^{|\mathcal{S}|} \mathbf{1}\left(\bm{\pi^{*}}(s) \neq \hat{\bm{\pi}}(s)\right).
\end{align}

We analyze the APE performance of Algorithm \ref{Algorithm: ensemble_link_learning} in comparison to the traditional $Q$-learning running on the original Markovian environment $\mathcal{M}^{(1)}$ and four different SMEs $\mathcal{M}^{(n)}$ for $n = \{2,3,4,5\}$ in Fig.\ref{Fig:sharp_decrease}, where $Q^{it}$ represents the APE of Algorithm \ref{Algorithm: ensemble_link_learning}. The simulations are conducted using the MIMO network with interference channels with a network size of 40000 using the following parameter values: $K=5$, $u_t=0.5$, $v=50$, $l=15$, $\gamma=0.99$, $\alpha_t=\frac{1}{1+\frac{t}{1000}}$, $\epsilon_t=\min(0.99^t, 0.01)$, $\lambda=1$, where $\lambda_n$ for $n \in \{1,2,3,4,5\}$ are estimated numerically. The parameters are optimized through cross-validation (see \cite{Ln_journal_supplementary_material} for details). The simulation is carried out 50 times, and the APE results are averaged. Clearly, a near-zero APE can be achieved with a significantly small number of iterations (less than 0.1 APE within 500 iterations), as can be seen in Fig.\ref{Fig:sharp_decrease}. The sharp decline in the $Q^{it}$ curve at the initial stages corresponds to the \textit{exploration} phase, followed by the \textit{exploitation} phase. Compared to other $Q$-learning algorithms, the exploration stage in Algorithm \ref{Algorithm: ensemble_link_learning} is significantly fast, highlighting the advantages of leveraging multiple SMEs to enhance exploration capabilities. Furthermore, the APE results demonstrate a non-monotonic pattern across various $n$ values, with the original environment $\mathcal{M}^{(1)}$ consistently achieving the smallest APE, as predicted by Proposition \ref{Proposition-4}.

\begin{table}[t]
\centering
\scriptsize
\setlength{\tabcolsep}{2pt}
\begin{tabular}{|c|c|c|c|}
 \hline
 Algorithm & Objective & \multicolumn{2}{c|}{Strategy} \\ [0.5ex]
 \cline{3-4}
 & & Estimator & Environment \\ [0.5ex]
 \hline\hline
 Simple Q (Q) \cite{bertsekas_book} & - & Single & Single \\ [0.5ex]
 \hline
 Speedy Q (SQ) \cite{speedy_q} & Convergence rate & Single & Single\\ [0.5ex]
 \hline
 Double Q (DQ) \cite{double-q} & Bias & Multi & Single\\ [0.5ex]
 \hline
 MaxMin Q (MMQ) \cite{maxmin-q} & Bias \& variance & Multi & Single\\ [0.5ex]
 \hline
 Ensemble Bootst. Q (EBQ) \cite{ensemble_bootstrap_q} & Bias & Multi & Single\\ [0.5ex]
 \hline
 Averaged DQN (ADQN) \cite{averaged_dqn} & Stability, Variance & Multi & Single\\ [0.5ex]
\hline
\textbf{Ensemble Synthetic Q (ESQL)} & \textbf{Variance, Learning speed} & \textbf{Multi} & \textbf{Multi} \\ [0.5ex]
\hline
\end{tabular}
\caption{$Q$-learning algorithm and variants}
\label{Table:Q-learning-variants}
\end{table}

For comparison, we consider several value-based model-free reinforcement learning algorithms with different objectives and implementation strategies (number of estimators and Markovian environments). To ensure a fair comparison, all algorithms follow the same strategy as Algorithm \ref{Algorithm: ensemble_link_learning}. Table \ref{Table:Q-learning-variants} provides an overview of these algorithms. For specific details on the parameter optimization of each algorithm, see \cite{Ln_journal_supplementary_material}.

\begin{figure}[t]
    \centering
    \subfloat[$Q$-learning algorithms for Model-1 \label{fig:ape_model1}]{{\includegraphics[width=4.4cm]{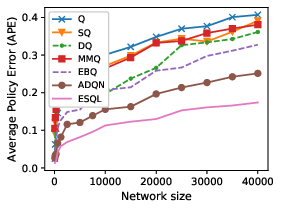}}}
    \subfloat[$Q$-learning algorithms for Model-2 \label{fig:ape_model2} ]{{\includegraphics[width=4.4cm]{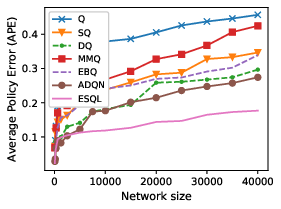}}}
 
    \subfloat[$Q$-learning algorithms for Model-3 \label{fig:ape_model3}]
    {{\includegraphics[width=4.4cm]{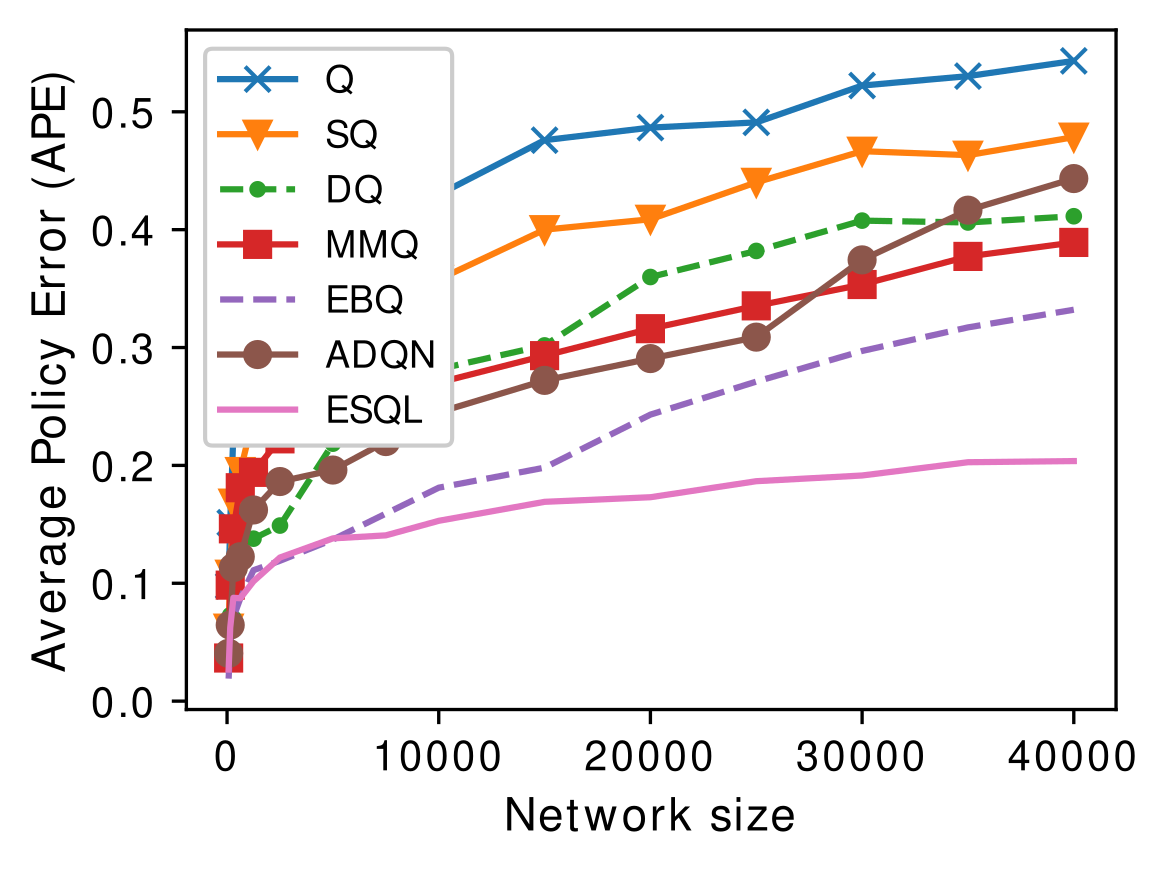}}}
    \subfloat[$Q$-learning algorithms for Model-4 \label{fig:ape_model4} ]{{\includegraphics[width=4.4cm]{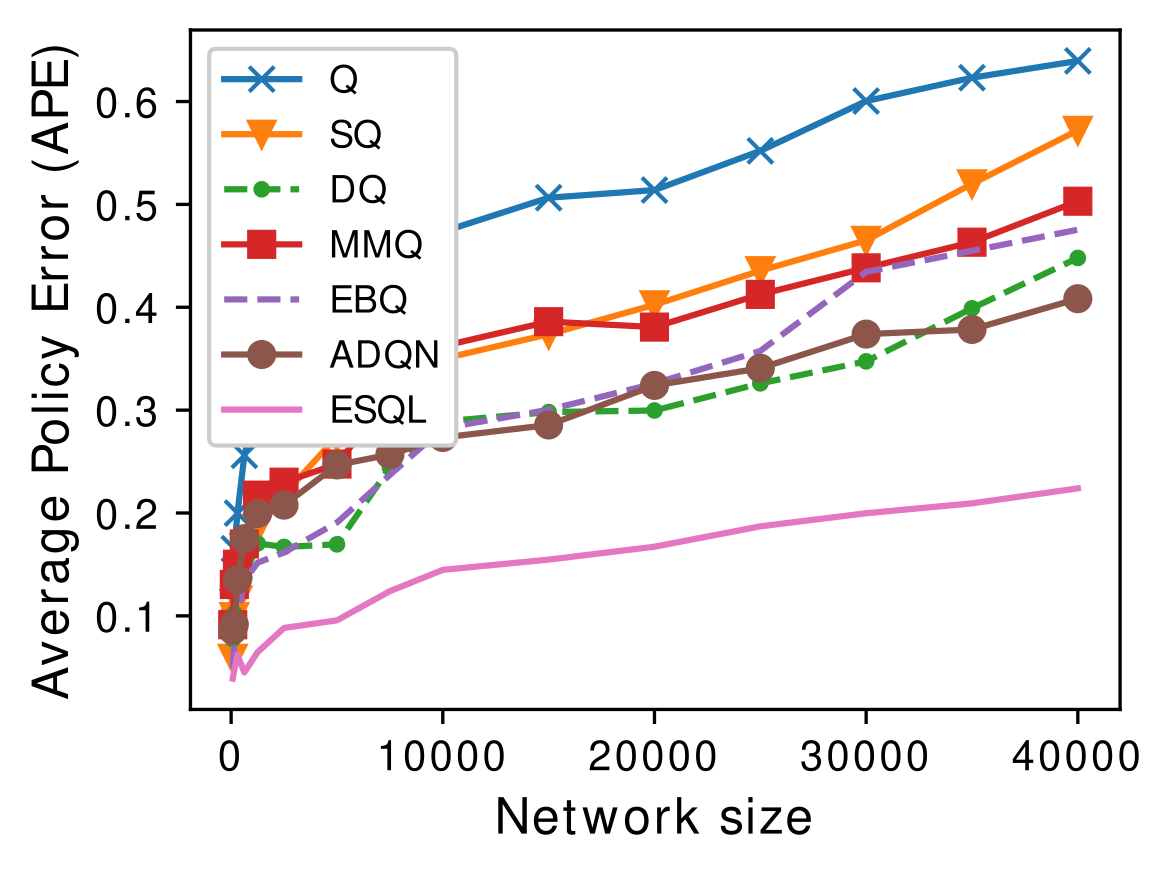}}}

    \subfloat[Non-Q-Learning algorithms for Model 3\label{Fig:ape_non_q_learning_algorithms} ]{{\includegraphics[width=4.4cm]{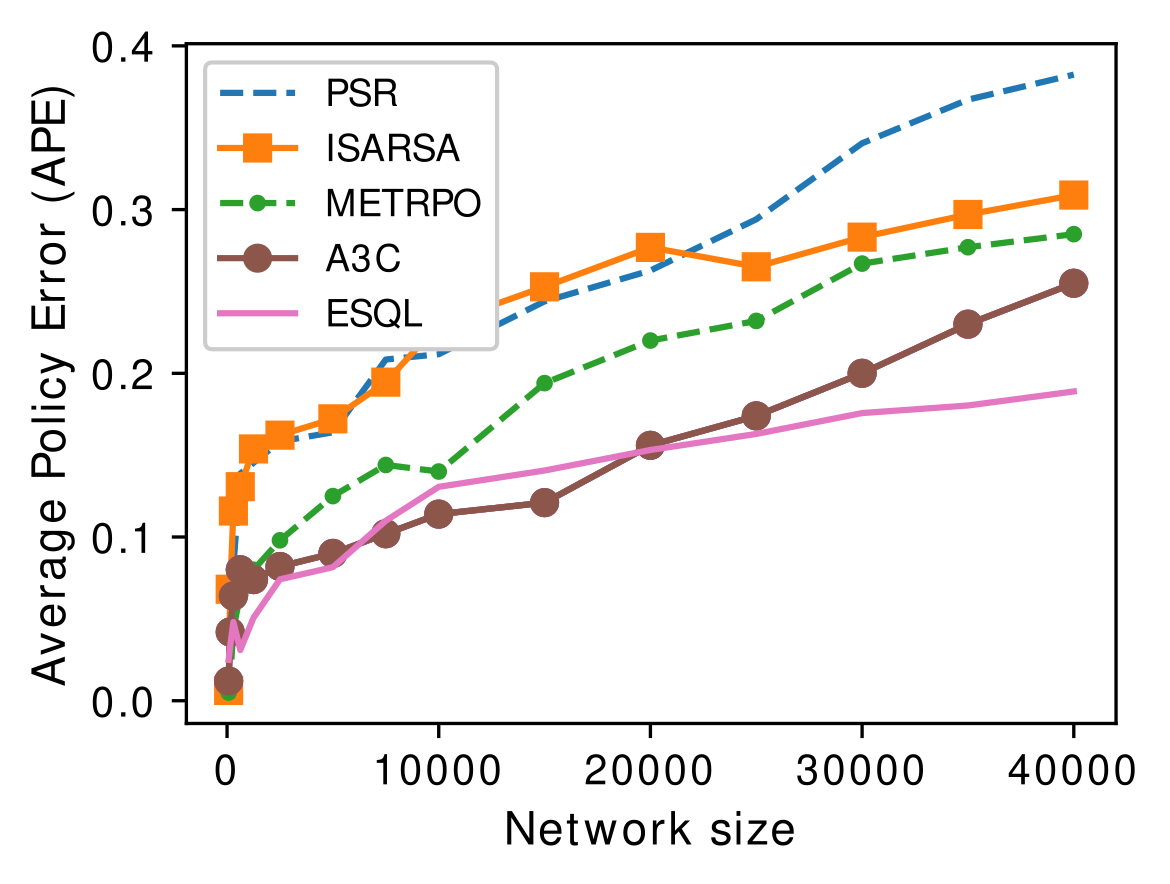}}}
    \caption{APE of different algorithms across different models}
\end{figure}

The APE of different algorithms as a function of network size across different network models are given in Fig.\ref{fig:ape_model1}-\ref{fig:ape_model4}. The proposed algorithm consistently outperforms other algorithms, achieving a lower APE across all models. In particular, it achieves 30\% less APE for model-1, 40\% less APE for model-2, 45\% less APE for model-3, and 50\% less APE for model-4. This performance improvement can be attributed to several reasons. Firstly, the utilization of multiple Markovian environments allows for deep and efficient exploration by aggregating higher-order relationships between states into a single estimate. Secondly, the adaptive weighting mechanism, based on policy comparison, exploits the most informative environments during training by assigning higher weights, unlike ensemble methods such as MMQ, DQ, or EBQ. Lastly, the algorithm leverages structural similarities among multiple Markovian environments, providing an advantage over neural network-based algorithms like ADQN that do not utilize such properties. The APE gains become larger for larger networks, which highlights the practicality of the algorithm in large-scale wireless networks. 

We also compare the performance of the proposed algorithm with the algorithms presented in our prior work \cite{talha_eusipco, talha_Pn_journal, talha_icassp} where the co-link representations are not used, and the cost-metric is the divergence between $Q$-functions. To ensure a fair comparison, we simulate each algorithm on the MIMO wireless network with interference channels with a network size of 40000. The parameters in Section \ref{subsec: APE_results} are employed for the proposed algorithm. For the other algorithms, we individually optimize the hyperparameters by cross-validation. Extensive simulations demonstrate that the proposed algorithm achieves a 15\% less APE (with 10\% less runtime) compared to \cite{talha_eusipco}, a 15\% less APE (with 20\% less runtime) compared to \cite{talha_icassp}, and a 10\% less APE (with 10\% less runtime) compared to \cite{talha_Pn_journal}. Furthermore, we observe that varying network parameters, such as arrival probability, buffer size, and the number of transmitters and receivers within a reasonable range, lead to a 25\% change in APE for \cite{talha_eusipco}, a 20\% change in APE for \cite{talha_icassp}, a 20\% change in APE for \cite{talha_Pn_journal}, and only a 5\% change in APE for the proposed algorithm, which indicates that the proposed algorithm is the most robust amongst our prior approaches. These results highlight that the proposed algorithm is particularly well-suited for optimizing wireless networks due to their unique structural properties \cite{talha_colink_journal}. 

We also compare to \emph{non-Q-learning RL} algorithms: (i) Value function-based policy sampling and reconstruction (PSR) \cite{liu2020sampled}, (ii) Improved Sarsa (ISARSA) \cite{improved_sarsa}, (iii) Model-ensemble trust-region policy optimization (METRPO) \cite{kurutach2018model}, and (iv) Asynchronous Advantage Actor Critic (A3C) \cite{mnih2016asynchronous}. The simulations are carried out using the same settings in Section \ref{subsec: APE_results}. See \cite{Ln_journal_supplementary_material} for further details.

ESQL achieves up to 25 \% less APE than other algorithms across large networks, as demonstrated in Fig.\ref{Fig:ape_non_q_learning_algorithms}.  PSR faces challenges in accurately estimating the value function, sensitivity to network parameters, and the need for near-perfect estimation of the PTT $\mathbf{\hat{P}}$. Although METRPO utilizes an ensemble of deep neural networks, highlighting the power of ensemble learning, it does not leverage the structural properties of multiple Markovian environments and faces challenges in generalization across large networks.  A3C produces a similar performance to ESQL (even slightly outperforms it for small network sizes) as both methods employ parallel learning. While there is no clear performance difference for small and modest-sized networks, ESQL achieves up to 25\% less APE for large networks because A3C suffers from the limitations of digital twins, the challenges to achieve fine-tuned training and accurate generalization for large networks, and the lack of utilization of the structure of the underlying network. Finally, as will be seen, A3C achieves its performance at the expense of high computational complexity.

\subsection{Average Runtime Complexity Performance}\label{subsec: runtime_results}

The average runtime complexity of Algorithm \ref{Algorithm: ensemble_link_learning} can be shown to be $O\left(\frac{|\mathcal{S}||\mathcal{A}| v}{K}f(l,\epsilon)\right)$, where $f$ is a non-monotonic function of $l$ and $\epsilon$. The proof of the runtime complexity of algorithms presented in \cite{talha_eusipco, talha_icassp} is applicable here. The runtime complexity increases with the network size ($|\mathcal{S}|$ and $|\mathcal{A}|$) and the number of visits to each state-action pair ($v$). However, the non-monotonicity of the function $f$ with respect to $l$ and $\epsilon$ suggests the existence of optimal values for these parameters, necessitating parameter-tuning. The complexity is also inversely proportional to the number of Markovian environments ($K$), which may seem counter-intuitive. This result follows from the fact that the number of samples that need to be collected from each Markovian environment decreases with $K$ \cite{talha_eusipco, talha_icassp}.

\begin{figure}[t]
    \centering
    \subfloat[Q-learning algorithms \label{Fig:time_different_algorithms}]{{\includegraphics[width=4.4cm]{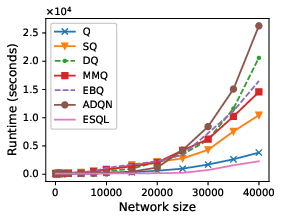}}}
    \hspace{-8pt}
    \subfloat[Non-Q-learning algorithms\label{Fig:time_other_algorithms} ]{{\includegraphics[width=4.5cm]{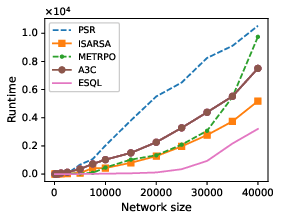}}}
    \caption{Runtime of different algorithms}
\end{figure}

The runtime of the proposed algorithm consists of three main components: sampling (ensuring each state-action pair is visited at least $v$ times), constructing multiple SMEs $K-1$ times using (\ref{Equ: sum_expression_for_Ln}), and the time until the slowest $Q$-learning algorithm on different SMEs converge (as they run in parallel). For other algorithms, their runtime represents the time until convergence. The runtimes of algorithms are consistently comparable across fixed network sizes and different network models. Thus, a single runtime result is presented in Fig.\ref{Fig:time_different_algorithms}. The proposed algorithm achieves 40\% less runtime than the other algorithms across large state-spaces as it utilizes higher-order SMEs to capture distant node relationships, reducing the need for long trajectories that are required for small-order SMEs. In addition, the algorithm improves exploration by running multiple Markovian environments simultaneously. This result is consistent with both Fig.\ref{Fig:sharp_decrease} and the theoretical average runtime complexity.

We also present the runtime complexity of our algorithm compared to non-Q-learning algorithms (in Section \ref{subsec: APE_results}) in Fig.\ref{Fig:time_other_algorithms}. Overall, the proposed algorithm achieves up to 40\% less runtime complexity than other algorithms across large networks. ISARSA is a simple algorithm and thus has relatively lower complexity than the other algorithms. In contrast, PSR exhibits high complexity due to increased sampling to obtain an accurate estimate of PTT and policy interpolation, and neural-network-based approaches (A3C and METRPO) suffer from long training for multiple networks.

We underline that as the algorithm complexity increases (e.g., Q $\rightarrow$ SQ $\rightarrow$ ADQN), the corresponding APE generally decreases, but the runtime complexity increases. However, the proposed algorithm achieves a small APE with a small runtime, overcoming the performance-complexity trade-off. We also see that the complexity reduction is independent of the network model, making the proposed algorithm an efficient approach for learning various complex environments.

\subsection{Memory complexity analysis}\label{subsec: aggregation_results}
Memory complexity is a critical concern in our algorithm, as we need to store $Q$-functions for $K$ different environments. This issue is a common drawback in similar algorithms \cite{double-q, maxmin-q, ensemble_bootstrap_q}, yet it is exacerbated here as a result of $K$ different environments. To address this concern, we introduced a state-action aggregation method in our previous work \cite{talha_eusipco}. This approach is based on the intuition that, under a smooth cost function, the $Q$-functions of neighboring states, for a given action, exhibit minimal differences. In other words, if the changes in the cost function for neighboring states are bounded by a small constant, we can represent the $Q$-functions of the $k$ nearest neighboring states (including the state itself) using a single $Q$-function. Choosing a large value for $k$ minimizes memory requirements and runtime complexity (since the size of state space is reduced) but comes at the cost of information loss, leading to a larger APE. Thus, a trade-off exists between the APE, and memory needs as well as runtime complexity.

\begin{figure}[t]
    \centering
    \includegraphics[width=0.3\textwidth]{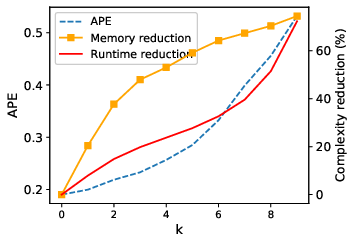}
    \caption{APE and memory reduction through state aggregation vs k}
    \label{Fig:ape_vs_memory_reduction}
\end{figure}

The simulations are carried out using the same settings in Section \ref{subsec: APE_results}. (One can observe that the cost function of the network model meets the specified condition.) The results are given in Fig.\ref{Fig:ape_vs_memory_reduction}, with the y-axes showing the APE and the percentage reduction in complexity (memory and runtime), respectively, while the x-axis represents the number of nearest neighbors $k$. For clarity, the case $k=0$ implies no aggregation. While the APE increases with $k$, the reduction in the memory needs and runtime complexity also increases. Thus, by appropriately choosing $k$, a good performance-complexity balance can be achieved. We also emphasize that the proposed state-action aggregation idea can similarly minimize the memory requirements for all four wireless network models.

\begin{figure}[t]
    \tiny
    \centering
    \subfloat[$|\Delta_{t}^{it}(s,a) - \Delta_{t-1}^{it}(s,a)|$ averaged overall $(s,a)$ vs iterations \label{Fig:proposition1}]
    {{\includegraphics[width=4.2cm]{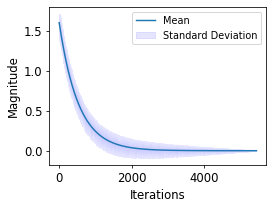}}}\hspace{0.2cm}
    \subfloat[$|\Delta_{t}^{it}(7,1)|$ vs iterations \label{fig:proposition23} ]{{\includegraphics[width=4.2cm]{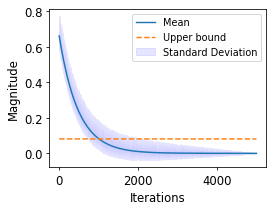}}}

    \subfloat[Weights across iterations \label{fig:weights} ]{{\includegraphics[width=4.15cm]{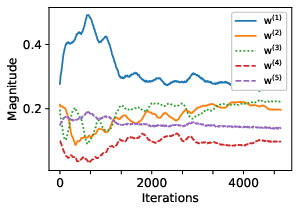}}}
    \subfloat[Expectation of error \label{fig:expectation} ]{{\includegraphics[width=4.3cm]{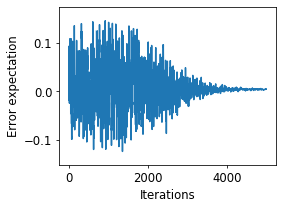}}}
 
    \subfloat[Upper bounds on error variance \label{fig:upper_bounds}]
    {{\includegraphics[width=4.25cm]{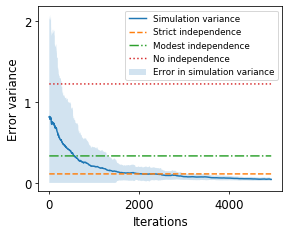}}}
    \subfloat[Averaged distance correlation \label{fig:adc}]
    {{\includegraphics[width=4.2cm]{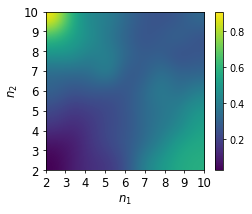}}}    
    
    \subfloat[Normal fit to $\mathcal{X}^{(n)}_t(7,1)$ for different $n$  \label{fig:q_func_error_all_n}]
    {{\includegraphics[width=4.3cm]{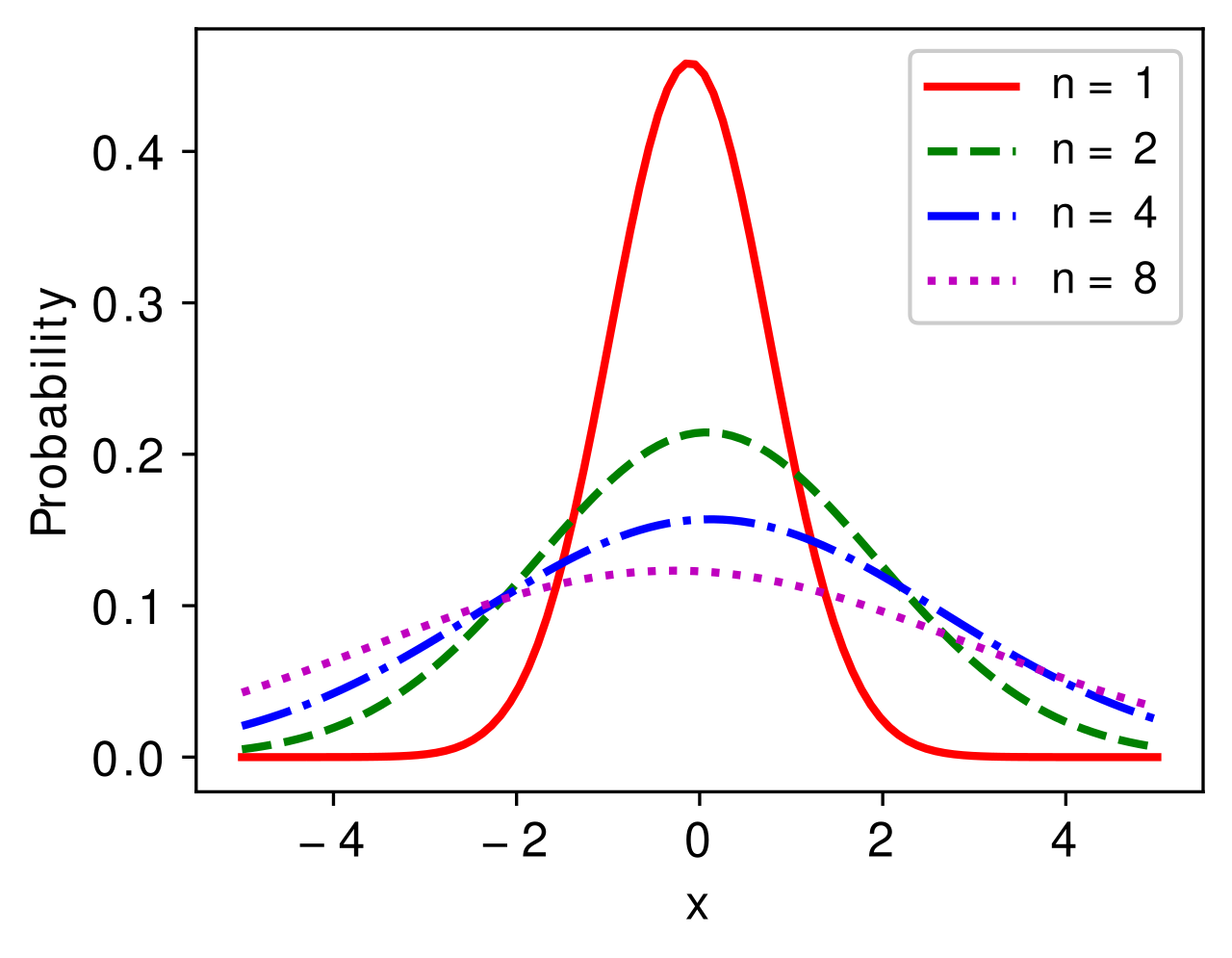}}} 
    \caption{Simulation of theoretical results}
\end{figure}

\subsection{Numerical consistency of propositions}

In this section, we simulate the results in the propositions and compare theoretical results with simulation results. The same simulation settings in Section \ref{subsec: APE_results} are employed.

The magnitude of the difference between the consecutive $Q$-function updates for the state-action pair $(s,a)$ averaged over all state-action pairs over time is shown in Fig.\ref{Fig:proposition1}, where the blue curve and shaded area represent the mean and standard deviation over 50 simulations. The magnitude consistently converges to zero with increasing iterations, as shown by Proposition \ref{Proposition-1}, which shows that Algorithm \ref{Algorithm: ensemble_link_learning} produces increasingly stable updates over time. We note that similar curves with different initial magnitudes and decay rates can also be shown for different $(s,a)$ pairs.

The magnitude of the $Q$-function updates for the state-action pair $(s,a) = (7,1)$ over time, $|\Delta_{t}^{it}(7,1)|$, is depicted in Figure \ref{fig:proposition23}. The blue curve and shaded area represent the mean and standard deviation over 50 simulations. Additionally, the dotted curve illustrates the upper bound on $|\Delta_{t}^{it}(7,1)|$ from Proposition \ref{Proposition-2}. Clearly, the magnitude of the updates in the limit can be upper bounded. The tightness of the upper bound depends on the state-action pair and the system parameters (including $K$). Furthermore, the result shows that the magnitude of the updates converges to zero as the iterations progress, thereby confirming the convergence of Algorithm \ref{Algorithm: ensemble_link_learning} as stated in Proposition \ref{Proposition-3}. To further validate the convergence result and practicality of Proposition \ref{Proposition-3}, we observe that the constants $\phi^{(n)}(7,1)$ for $n = 1,2,3,4,5$ exist and are all less than 1. (In particular, the values are 0.998, 0.9925, 0.9898, 0.9946, and 0.9808, respectively.)

The weights of five different Markovian environments ($\mathcal{M}^{(n)}$ for $n=1,2,3,4,5$) over time are illustrated in Fig.\ref{fig:weights}. Initially, there is a sharp increase in the weight $\mathbf{w}^{(1)}$ because $\mathcal{M}^{(1)}$ is the original environment, and there are insufficient samples to capture the higher-order relationships. Additionally, it is unclear which $\mathcal{M}^{(n)}$ provides the most useful samples, as the weights $\mathbf{w}^{(n)}$ for $n > 1$ continue to fluctuate. Moreover, it is not clear which $\mathcal{M}^{(n)}$ provides the most useful samples as the weights $\mathbf{w}^{(n)}$ for $n > 1$ keep changing. As the iterations continue, the weight $\mathbf{w}^{(1)}$ gradually decreases up to a certain point, but $\mathcal{M}^{(1)}$ remains the most useful environment. On the other hand, the weights $\mathbf{w}^{(n)}$ for $n > 1$ increase and eventually converge to a fixed value. The final magnitudes of these weights, however, exhibit a non-monotonic pattern across $n$, and can be shown to follow the compact expression in Proposition \ref{Proposition-4}. Similar patterns can be observed across different networks and models, although there may be variations in (i) the final weight values, (ii) the order of environment utilities, and (iii) the iteration index at which the weights converge.

We approximate the expectation of the $Q$-function errors numerically as follows:

\begin{align}
    \mathbb{E}[\mathcal{E}_t(s,a)] \approx \frac{1}{2\Delta_t}\sum_{t' = t - \Delta_t}^{t + \Delta_t} \mathcal{E}_t(s,a),
\end{align}
with $\Delta_t \ll t$. The expectation of the $Q$-function error for the state-action pair $(s,a)=(7,1)$, $\mathbb{E}[\mathcal{E}_t(7,1)]$ over time, in Proposition \ref{Proposition-5} is shown in Fig.\ref{fig:expectation}, where the result is averaged over 50 simulations. Clearly, the expectation of the $Q$-function errors converges to zero with iterations. We approximate the variance of the $Q$-function errors similarly as follows:
\begin{align}
    {\mathbb{V}[\mathcal{E}_t(s,a)] \approx \frac{1}{2\Delta_t}\sum_{t' = t - \Delta_t}^{t + \Delta_t} \mathcal{E}_t(s,a)^2 \text{--} \Big[\frac{1}{2\Delta_t}\sum_{t' = t - \Delta_t}^{t + \Delta_t} \mathcal{E}_t(s,a)\Big]^2},
\end{align}
with $\Delta_t \ll t$. The three upper bounds on the error variance based on different independence assumptions from Proposition \ref{Proposition-1} and the simulation variance for $(s,a)=(7,1)$ (with $\Delta_t = 20$) are shown in Fig.\ref{fig:upper_bounds}, where the blue curve and shaded area represents the mean and standard deviation over 50 simulations. Clearly, a more strict independence assumption leads to a tighter upper bound. Moreover, as iterations continue, the simulation variance becomes smaller than all upper bounds and eventually converges to zero, which is in line with Fig.\ref{Fig:proposition1} and Fig.\ref{fig:proposition23}. Herein, the choice of $\lambda$ and $u$ may change the initial error variance and its decay rate. These two results (the expectation and variance converging to zero) show that the distribution assumption (\ref{Equ: distribution_assumption}) is accurate. Similarly, Fig.\ref{fig:q_func_error_all_n} demonstrates that the $Q$-function errors of different environments, across various values of $n$, can be well-modeled by zero-mean normal distributions with different variances, which further validates the assumption (\ref{Equ: distribution_assumption}).

To assess the practicality of Proposition \ref{Proposition-5} and independence assumptions, we employ the averaged distance correlation (ADC) metric \cite{distance_correlation_ref}. In particular, we compute the ADC between $\mathcal{X}^{(n_1)}_{t_1}(7,1)$ and $\mathcal{X}^{(n_2)}_{t_2}(7,1)$ for $n_1,n_2 \in [2,10]$ and average the results over all $t_1 \neq t_2 \in [0,5000]$. The ADC captures both linear and non-linear correlations without assuming specific variable distributions like Pearson's correlation coefficient. The results are presented in Fig.\ref{fig:adc}. For all $n_1=n_2$ (right diagonal), the ADC is very close to 0, indicating independent $Q$-function errors across different times, validating the modest independence assumption. When $n_1$ and $n_2$ are close, the ADC is almost 0, also confirming the strict independence assumption. However, for larger and distant $n_1$ and $n_2$, weak correlations may emerge. Thus, the strict independence assumption may not always hold contrary to the modest independence assumption.

\subsection{Optimization of the order $n$}\label{subsection:optimization_n}

Determining the optimal set of environments involves determining the optimal number of environments as well as their identity (e.g. for $K=4$,  $\{1^{st},2^{nd},3^{rd},4^{th}\}$ versus $\{1^{st},2^{nd},3^{rd},5^{th}\}$ versus all other collections of four possible environments). We offer a strategy in \cite{talha_Pn_journal} for arbitrary graphs/networks, whereas herein, we utilize an approach from \cite{talha_colink_journal} where we proposed the colink representation for wireless network policy optimization. In particular, we adapt the classical definition of the Bellman error (BE) (see \cite{bellman_error, bellman_error_2}) to the difference between the $Q$-functions of the original environment and $n^{th}$ environment under the policy $\pi$ as:
\begin{align}\label{Equ:bellman_error}
BE(\mathbf{Q}_{\boldsymbol{\pi}}^{(n)})=\mathbf{c}_{\boldsymbol{\pi}}+\gamma \mathbf{P}_{\boldsymbol{\pi}} \mathbf{Q}_{\boldsymbol{\pi}}^{(n)}-\mathbf{Q}_{\boldsymbol{\pi}}^{(1)}.
\end{align}

In \cite{talha_colink_journal}, the true PTT is known; herein, we apply the algorithm presented in \cite{talha_colink_journal} to the estimated PTT ($\mathbf{\hat{P}}$) to obtain the approximate policy to be used in (\ref{Equ:bellman_error}). We then select the top $K$ orders (environments) that minimize the $l_2$ norm $\|BE(\mathbf{Q}_{\boldsymbol{\pi}}^{(n)})\|_2$.

\begin{figure}[t]
    \centering
    \subfloat[Bellman error vs order $n$ \label{fig:bellman_error_order}]{{\includegraphics[width=4.4cm]{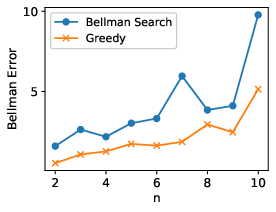}}}
    \subfloat[APE change vs order $n$ \label{fig:bellman_error_ape} ]{{\includegraphics[width=4.4cm]{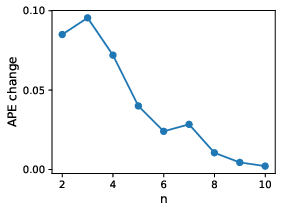}}}
    \caption{The effect of the order $n$ on performance}
\end{figure}

We present the Bellman error across $n \in [2,10]$ in Fig.\ref{fig:bellman_error_order} using two methods: (i) Bellman search (blue curve) using \cite{talha_colink_journal} and (\ref{Equ:bellman_error}) and (ii) Greedy method (orange curve) that exhaustively searches the environment and policy that minimize the APE of Algorithm \ref{Algorithm: ensemble_link_learning}. The same simulation settings in Section \ref{subsec: APE_results} are employed. The Bellman error clearly exhibits a non-monotonic behavior with $n$, but small values of $n$ generally produce smaller errors. While the blue curve consistently shows higher Bellman errors for all $n$ (since it uses an approximate policy), the relative ranking of orders is similar across the two methods, which suggests that the Bellman search can effectively rank environments. For $K=3$, Bellman search provides an ordered set \{2,4,3\} while the greedy approach selects \{2,3,4\}. Thus, both methods find the same set of environments, but the relative informativeness for environments $3$ and $4$ are flipped. Similar trends apply to different $K$ values. Overall, the Bellman search achieves 94\% accuracy in finding the optimal $K$ environments (for $K \in [2,10]$) and has 70\% less runtime complexity than the greedy method, hence enabling an effective initialization of Algorithm \ref{Algorithm: ensemble_link_learning}, which is particularly useful for real-world wireless applications with time, memory or resource constraints. 

The sensitivity of Algorithm \ref{Algorithm: ensemble_link_learning} with respect to order $n$ is shown in Fig.\ref{fig:bellman_error_ape}. In particular, we initialize Algorithm \ref{Algorithm: ensemble_link_learning} with the first ten Markovian environments ($K=10$). Then, we run the algorithm with $K=9$ after excluding the $n^{th}$ environment for $n \in [2,10]$ and report the absolute change in APE. While APE changes vary with $n$, the maximum change in APE is under 10\%, highlighting the robustness of Algorithm \ref{Algorithm: ensemble_link_learning} with respect to $n$. The result also closely follows Fig.\ref{fig:bellman_error_order} as the set of orders that lead to the largest change in APE are the same as the orders that minimize the Bellman error. Similar results can also be shown for different network settings.

\subsection{Practical implications and interpretability}\label{subsection:practical_implications}

By integrating the novel concept of digital cousins into ensemble $Q$-learning, the proposed algorithm effectively accounts for the influence of actions and states that span \textbf{multiple hops} in the wireless network graph. This addresses scenarios like multiple packet changes in transmitter buffers, significant channel quality variations, or multi-step movements of mobile transmitters. This could enhance the performance, reliability, and robustness of real-world wireless networks.

When there is a high probability of multiple data arrivals within a short period, considering buffer occupancy changes beyond a single data packet ($n^{th}$ digital cousin (environment) considers $n$ packet changes at a time) helps transmitters proactively adjust their transmission strategies to avoid buffer congestion, leading to more stable buffer occupancy levels. If there is a high probability of consecutive data losses, transmitters can dynamically adjust the transmission rate to maintain a reliable communication link. This can improve overall network robustness, particularly for densely populated networks with many transmitters and receivers. On the other hand, when there is a high probability of the channel deteriorating in the next time steps, transmitters must avoid making sub-optimal decisions. Without considering $n$-hop transitions, transmitters might assume the good channel conditions will persist and continue transmitting data continuously. However, by incorporating $n$-hop transition information, transmitters can intelligently anticipate the potential deterioration in the channel in the future and adjust their transmission rate to avoid unnecessary energy consumption. Similarly, more balanced traffic loads at receivers can be achieved via traffic smoothing by considering $n$-step movements and future locations of transmitters in wireless networks with mobile transmitters.

\subsection{Open Questions}\label{Subsection:limitations}

The proposed algorithm provides a novel implementation of digital cousins via the colink method and enables reductions in sample and runtime complexities as well as performance improvements over the consideration of general graphs \cite{talha_Pn_journal}. As noted previously, one open question is designing a modest complexity strategy for determining the optimal number of environments ($K$) to employ in Algorithm \ref{Algorithm: ensemble_link_learning}. While more environments improve performance and reduce runtime complexity, there are diminishing returns as seen \cite{talha_Pn_journal}. ESQL inherits several properties from traditional $Q$-learning and hence is particularly tailored to finite but large discrete state-action spaces. Thus, another open question is addressing continuous state-action spaces; thus discretization or approximation via neural networks is an option. However, we see that given the structural properties of wireless networks, a pure non-parametric approach such as a neural network also has its own limitations. The quality of sampling and the accuracy of the estimated PTT also impacts performance as discussed in Section \ref{subsec: digital_cousins}; whether further improvements are possible, given that our current strategy explicitly exploits the structure in the PTT and associated colink matrices, is unclear. There are also design issues involved in determining an appropriate normalization strategy to form the digital cousins, as colink matrices are not necessarily row-stochastic. Hence, one can consider alternative methods such as $\ell_1$, $\ell_2$, linear, softmax, min-max, etc).

\section{Conclusions}\label{sec: conclusion}
We herein presented a novel $Q$-learning algorithm to mitigate the performance and complexity challenges of the traditional $Q$-learning to solve policy optimization for real-world wireless networks by leveraging the novel concept of \textit{digital cousins}.  Digital cousins are a collection of distinct but structurally related synthetic Markovian environments that offer strong improvements over traditional digital twins. The current work synthesizes the notion of digital cousins with low complexity, high-performance modeling of wireless networks via co-link representations. The proposed algorithm employs multiple $Q$-function estimators on these environments and fuses the outputs into a single estimate based on an adaptive weighting mechanism. Simulations across a variety of representative wireless networks show that the proposed algorithm produces a near-optimal policy with significantly lower complexity and outperforms the existing $Q$-learning and several state-of-the-art RL algorithms and prior work in terms of accuracy, runtime complexity, and robustness. The stability and convergence of the algorithm are rigorously analyzed from deterministic and probabilistic viewpoints, and several theoretical upper bounds on the first and second-order statistics of the $Q$-function errors are given. It is also shown that the simulation results closely follow the theoretical results. 

\appendix\label{sec: appendix}
\subsection{Proof of Proposition 1}\label{Appendix: proposition_1}
The following expressions are valid for all $(s,a)$ pairs; hence, we drop the $(s,a)$ notation for simplicity. The $Q$-function output of Algorithm \ref{Algorithm: ensemble_link_learning} can be expressed as follows:
\begin{align}
    \mathbf{Q}_{t}^{it} &= (1-u)\sum_{i=0}^{t-1}u^{t-i-1}\sum_{n=1}^{K}\mathbf{w}_i^{(n)}\mathbf{Q}_i^{(n)}\label{Equ:Q(t+1)_it_expression},
\end{align}
which can be obtained by repeatedly plugging the expression of $\mathbf{Q}_{t-1}^{it}$ into the expression of $\mathbf{Q}_{t}^{it}$ in line 11 in Algorithm \ref{Algorithm: ensemble_link_learning} for all $t$. Using (\ref{Equ:Q(t+1)_it_expression}) and algebraic manipulations, the following can be shown:
\begin{align}
    \mathbf{Q}_{t}^{it} - u \mathbf{Q}_{t-1}^{it} = (1-u)\sum_{n=1}^{K}\mathbf{w}_{t-1}^{(n)}\mathbf{Q}_{t-1}^{(n)}\label{Equ:prop1_1}.
\end{align}

Let $\Delta_{t}^{it} = \mathbf{Q}_{t+1}^{\mathbf{it}} - \mathbf{Q}_{t}^{\mathbf{it}}$ and $\epsilon_{t}^{(n)} = \mathbf{w}_{t+1}^{(n)}\mathbf{Q}_{t+1}^{(n)} - \mathbf{w}_{t}^{(n)}\mathbf{Q}_{t}^{(n)}$. If we rewrite (\ref{Equ:prop1_1}) with the variable change $t \rightarrow t+1$, and subtract (\ref{Equ:prop1_1}) from the new expression side by side, we obtain the following:
\begin{align}
    \Delta_{t}^{it}-u\Delta_{t-1}^{it} &= (1-u)\sum_{n=1}^{K}\Big[\mathbf{w}_{t}^{(n)}\mathbf{Q}_{t}^{(n)} - \mathbf{w}_{t-1}^{(n)}\mathbf{Q}_{t-1}^{(n)}\Big].\label{Equ:prop1_2}
\end{align}

Then, we can show the following:
\begin{align}
    \Delta_t^{it} &= u\Delta_{t-1}^{it} + (1-u)\sum_{n=1}^{K}\epsilon_{t-1}^{(n)}.\label{Equ:prop1_3} \\
     &< \Delta_{t-1}^{it} + (1-u)\sum_{n=1}^{K}\epsilon_{t-1}^{(n)}\label{Equ:prop1_4},
\end{align}
where (\ref{Equ:prop1_3}) follows from (\ref{Equ:prop1_2}) and the definition of $\epsilon_{t-1}^{(n)}$ and (\ref{Equ:prop1_4}) follows from the fact that $u \in (0,1)$. Then, using (\ref{Equ:prop1_4}) and the triangle inequality, we can bound the difference between consecutive updates as follows:
\begin{align}
    |\Delta_t^{it} - \Delta_{t-1}^{it}| &< (1-u)\sum_{n=1}^{K}|\epsilon_{t-1}^{(n)}|\label{Equ:prop1_5}.
\end{align}

By the convergence of $Q$-learning, $\lim_{t\rightarrow\infty}\mathbf{Q}_{t-1}^{(n)} = \mathbf{Q}_{t}^{(n)}$, and by the convergence of the weights from Proposition \ref{Proposition-4}, $\lim_{t\rightarrow\infty}\mathbf{w}_{t-1}^{(n)} = \mathbf{w}_{t}^{(n)}$ for all $n$. Thus, $\lim_{t\rightarrow\infty}|\epsilon_{t-1}^{(n)}| = 0$ for all $n$, and $\lim_{t\rightarrow\infty}|\Delta_t^{it} - \Delta_{t-1}^{it}| = 0$.

\subsection{Proof of Proposition 2}\label{Appendix: proposition_2}
If we express (\ref{Equ:prop1_2}) with the variable change $t \rightarrow t-1$, multiply by $u$, and then add to the expression (\ref{Equ:prop1_2}) side by side, we obtain the following:
\begin{align}
    \Delta_t^{it}-u^2\Delta_{t-2}^{it} &= (1-u)\sum_{n=1}^{K}\epsilon^{(n)}_{t-1}-u\epsilon^{(n)}_{t-2}\label{Equ:prop2_1}.
\end{align}

If we carry out the same operation for all $t$, we obtain the following:
\begin{align}
    \Delta_t^{it}-u^t\Delta_0^{it} &= (1-u)\sum_{k=0}^{t-1} u^k\sum_{n=1}^{K}\epsilon^{(n)}_{t-k-1}\label{Equ:prop2_2}.
\end{align}

Using the fact that $\Delta_0^{it} = 0$, we obtain the compact expression for $\Delta_t^{it}$ as follows:

\begin{align}
    \Delta_t^{it} &= (1-u)\sum_{k=0}^{t-1} u^k\sum_{n=1}^{K}\epsilon^{(n)}_{t-k-1}\label{Equ:prop2_3}.
\end{align}

Let $\theta^{(n)}$ be the smallest positive constant that satisfies $|\epsilon^{(n)}_{t}| \leq \theta^{(n)}$ for all $t,n$. Then, we proceed as follows:
\begin{align}
    |\Delta_t^{it}| &\leq (1-u)\sum_{k=0}^{t-1} u^k\sum_{n=1}^{K}\big|\epsilon^{(n)}_{t-k-1}\big|\label{Equ:prop2_4}.\\
    &\leq (1-u)\sum_{k=0}^{t-1} u^k\sum_{n=1}^{K}\theta^{(n)}\label{Equ:prop2_5}.\\
    &\leq \sum_{n=1}^{K}\theta^{(n)}(1-u^t),\label{Equ:prop2_6}
\end{align}
where (\ref{Equ:prop2_4}) follows by taking the absolute value of both sides in (\ref{Equ:prop2_3}) and using the triangle inequality, (\ref{Equ:prop2_5}) follows the definition of $\theta^{(n)}$, and (\ref{Equ:prop2_6}) follows from the sum of finite geometric series and the fact that $u \in (0,1)$. If we take the limit of both sides, using the fact that $\lim_{t\rightarrow\infty}u^t = 0$, the result follows.

\subsection{Proof of Corollary 1}\label{Appendix: corollary_1}

We want to find the time index $t$ at which the $|\Delta_t^{it}| \leq \beta$. If we convert (\ref{Equ:prop2_6}) into equality and solve for $t$ as follows, we obtain the desired result: $\beta = \sum_{n=1}^{K}\theta^{(n)}(1-u^t).$
%\begin{align}
 %   
%\end{align}

\subsection{Proof of Proposition 3}\label{Appendix: proposition_3}
If there exist constants $\phi^{(n)} \in (0,1)$ such that $\frac{|\epsilon^{(n)}_{t+1}|}{|\epsilon^{(n)}_{t}|} \leq \phi^{(n)}$ for all $n,t$, we choose $\phi = \max\limits_{n \in \{1,2,..K\}}\phi^{(n)}$ and proceed as:
\begin{align}
    |\Delta_t^{it}| &\leq (1-u)\sum_{k=0}^{t-1} u^k\sum_{n=1}^{K}\big|\epsilon^{(n)}_{t-k-1}\Big|\label{Equ:prop3_1}\\
    &\leq (1-u)\sum_{k=0}^{t-1} u^k\sum_{n=1}^{K}\big|\epsilon_0^{(n)}{\phi}^{t-k-1}\Big|\label{Equ:prop3_2}\\
    &\leq (1-u)\sum_{k=0}^{t-1} u^k\phi^{t-k-1}\Big|\sum_{n=1}^{K}\epsilon_0^{(n)}\Big|\label{Equ:prop3_3}\\
    &\leq (1-u)\sum_{k=0}^{t-1} \max(u,\phi)^{t-1}\Big|\sum_{n=1}^{K}\epsilon_0^{(n)}\Big|\label{Equ:prop3_4}\\
    &\leq (1-u)t\max(u,\phi)^{t-1}\Big|\sum_{n=1}^{K}\epsilon_0^{(n)}\Big|,\label{Equ:prop3_5}
\end{align}
where (\ref{Equ:prop3_1}) follows from (\ref{Equ:prop2_4}), (\ref{Equ:prop3_2}) follows from by repetitively plugging the inequality expression for $|\epsilon^{(n)}_{t-1}|$ ($|\epsilon^{(n)}_{t-1}|\leq\phi^{(n)}|\epsilon^{(n)}_{t-2}|$) into the expression for $|\epsilon^{(n)}_{t}|$ ($|\epsilon^{(n)}_{t}|\leq\phi^{(n)}|\epsilon^{(n)}_{t-1}|$) for all $t$, (\ref{Equ:prop3_3}) follows from the fact that $\phi^{(n)}$ is a positive constant, (\ref{Equ:prop3_4}) follows from by bounding the term by choosing the maximum of $u$ and $\phi$, and (\ref{Equ:prop3_5}) follows from the fact that $u$ and $\phi$ are independent of time index $k$. If we take the limit of both sides in (\ref{Equ:prop3_5}):
\begin{align}
    \lim_{t\rightarrow\infty}|\Delta_t^{it}| = 0,\label{Equ:prop3_6}
\end{align}
which follows from the fact that $u,\phi \in (0,1)$.

\subsection{Proof of Proposition 5}\label{Appendix: proposition_5}
We first prove the expectation.
\begin{align}
    \lim_{t\rightarrow \infty}\mathcal{E}_t &= \lim_{t\rightarrow \infty}\mathbf{Q}_{t}^{it} - \mathbf{Q}^{*}.\label{Equ:def_0}\\
    &= \lim_{t\rightarrow \infty}(1-u)\sum_{i=0}^{t-1}u^{t-i-1}\sum_{n=1}^K\mathbf{w}_i^{(n)}\mathbf{Q}_i^{(n)} - \mathbf{Q}^{*}.\label{Equ:def_1}\\
    &= \lim_{t\rightarrow \infty}(1-u)\sum_{i=0}^{t-1}u^{t-i-1}\sum_{n=1}^K\mathbf{w}_i^{(n)}(\mathbf{Q}_i^{(n)} - \mathbf{Q}^{*}).\label{Equ:def_2}\\
    &= \lim_{t\rightarrow \infty}(1-u)\sum_{i=0}^{t-1}u^{t-i-1}\sum_{n=1}^K\mathbf{w}_i^{(n)}\mathcal{X}^{(n)}_i, \label{Equ:def_2222}
\end{align}
where (\ref{Equ:def_1}) follows from (\ref{Equ:Q(t+1)_it_expression}), (\ref{Equ:def_2}) follows from the fact that $\sum_{n=1}^K\mathbf{w}_t^{(n)} = 1$ for all $t$, and $(1-u)\sum_{i=0}^{t-1}u^{t-i-1} = 1$ as $t \rightarrow \infty$, and (\ref{Equ:def_2222}) follows from (\ref{Equ: distribution_assumption}). If we take the expectations of both sides:
\begin{align}
    \lim_{t\rightarrow \infty}\mathbb{E}[\mathcal{E}_t] &= \lim_{t\rightarrow \infty}(1-u)\sum_{i=0}^{t-1}u^{t-i-1}\sum_{n=1}^K\mathbf{w}_i^{(n)}\mathbb{E}[\mathcal{X}^{(n)}_i] = 0\label{Equ:exp_1}
\end{align}
which follows from the linearity of expectation and (\ref{Equ: distribution_assumption}).

We now prove the upper bound on the variance for the \textit{modest independence} case.
$\lim_{t\rightarrow \infty}\mathbb{V}[\mathcal{E}_t] = $
\begin{align}
     &= \lim_{t\rightarrow \infty}\mathbb{V}\Big[(1-u)\sum_{i=0}^{t-1}u^{t-i-1}\sum_{n=1}^K\mathbf{w}_i^{(n)}\mathcal{X}^{(n)}_i\Big].\label{Equ:var_1}\\
    &= \lim_{t\rightarrow \infty}(1-u)^2\Big[\sum_{i=0}^{t-1}u^{2(t-i-1)}\big[ \sum_{n=1}^K{(\mathbf{w}_i^{(n)})}^2\mathbb{V}[\mathcal{X}^{(n)}_i] \nonumber \\ +& 2\sum_{n=1}^K\sum_{m=n+1}^K\mathbf{w}_i^{(n)}\mathbf{w}_i^{(m)}\operatorname{Cov}(\mathcal{X}^{(n)}_i, \mathcal{X}^{(m)}_i)\big]\Big]. \label{Equ:var_2}\\
    &\leq \lim_{t\rightarrow \infty}(1-u)^2\Big[\sum_{i=0}^{t-1}u^{2(t-i-1)}\big[ \sum_{n=1}^K{(\mathbf{w}_i^{(n)})}^2\mathbb{V}[\mathcal{X}^{(n)}_i] \nonumber \\ +& 2\sum_{n=1}^K\sum_{m= n+1}^K\mathbf{w}_i^{(n)}\mathbf{w}_i^{(m)}\sqrt{\mathbb{V}[\mathcal{X}^{(n)}_i]\mathbb{V}[\mathcal{X}^{(m)}_i}]\big]\Big].\label{Equ:var_4}\\
    &\leq \lim_{t\rightarrow \infty}(1-u)^2\Big[\sum_{i=0}^{t-1}u^{2(t-i-1)}\big[ \sum_{n=1}^K\mathbf{w}_i^{(n)}\mathbb{V}[\mathcal{X}^{(n)}_i] \nonumber \\ +& 2\sum_{n=1}^K\sum_{m=1}^K\mathbf{w}_i^{(n)}\mathbf{w}_i^{(m)}\sqrt{\mathbb{V}[\mathcal{X}^{(n)}_i]\mathbb{V}[\mathcal{X}^{(m)}_i}]\big]\Big].\label{Equ:var_5}\\
    &\leq \lim_{t\rightarrow \infty}(1-u)^2\Big[\sum_{i=0}^{t-1}u^{2(t-i-1)}\big[ \sum_{n=1}^K\mathbf{w}_i^{(n)}\frac{\lambda^2}{3} \nonumber \\ +& 2\sum_{n=1}^K\sum_{m=1}^K\mathbf{w}_i^{(n)}\mathbf{w}_i^{(m)}\frac{\lambda^2}{3}\big]\Big].\label{Equ:var_6}\\
    &\leq \lim_{t\rightarrow \infty}(1-u)^2\Big[\sum_{i=0}^{t-1}u^{2(t-i-1)}\lambda^2\Big]. \label{Equ:var_7}\\
    &\leq \frac{(1-u)}{(1+u)}\lambda^2,\label{Equ:var_8}
\end{align}
where (\ref{Equ:var_1}) follows by taking the variance of both sides in (\ref{Equ:def_2222}), (\ref{Equ:var_2}) follows from the properties of the variance operator and the modest independence assumption, (\ref{Equ:var_4}) follows from the Cauchy-Schwarz inequality for the variance, (\ref{Equ:var_5}) follows from the fact that $\mathbf{w}_t^{(n)} \leq 1$ and dropping the constraint in the second summation, (\ref{Equ:var_6}) follows from (\ref{Equ: distribution_assumption}) and $\lambda = \max\limits_n\lambda_n$, (\ref{Equ:var_7}) follows from the fact $\sum_{n=1}^K\mathbf{w}_t^{(n)} = 1$ for all $t$, and (\ref{Equ:var_8}) follows from the infinite geometric sum formula and $u \in (0,1)$.

If we assume \textit{strict independence}, the cross terms in (\ref{Equ:var_2}) to (\ref{Equ:var_6}) disappear, and the upper bound will be smaller by a factor of 3. Refer to \cite{Ln_journal_supplementary_material} for the proof of the upper bound on the variance for no independence case.

\bibliographystyle{unsrt}
\bibliography{references.bib}
\end{document}